%% file: arxiv_main.tex
\documentclass[11pt]{article}

\usepackage[margin=1in]{geometry}
\usepackage{graphicx}
\usepackage{booktabs}
\usepackage{amsmath}
\usepackage[hidelinks]{hyperref}
\usepackage{float}
\usepackage{xcolor}
\usepackage[normalem]{ulem}
\usepackage{cite}

\title{With the Right Representation, Classical Machine Learning Rivals Deep Models for Olfactory Mixture Similarity}
\author{Ze Dong\textsuperscript{1}, Yi Wang\textsuperscript{1}, and Ambuj Tewari\textsuperscript{1,*}\\[0.5em]
\normalsize\textsuperscript{1}Department of Statistics, University of Michigan, Ann Arbor, MI, United States\\
\normalsize\textsuperscript{*}Corresponding author: \texttt{tewaria@umich.edu}}
\date{}

\begin{document}
\maketitle

\begin{abstract}
Modeling olfactory mixtures remains a critical challenge in machine olfaction because the task requires reasoning over relationships between sets of odorants rather than over individual molecules in isolation. However, prior comparisons between deep and classical approaches also varied how the relationship between the two mixtures was represented, since the classical baselines primarily relied on weak representations that described each mixture independently and provided little direct signal about how the mixtures differed. In this work, we revisit classical learning using explicit symmetric comparison features on unordered pairs of mixtures, evaluating these representations with XGBoost and kernel ridge regression (KRR). We find that XGBoost with explicit comparison features achieves performance competitive with specialized deep mixture models on the main benchmark. Ablation studies show that coordinate-wise absolute differences alone provide strong predictive performance, while combining them with sums and products yields further improvements. KRR also benefits substantially from explicit comparison features, with further gains from appropriate kernel choice and preprocessing. Overall, these results suggest that explicitly encoding how two mixtures differ is an important ingredient for classical machine learning in low-data olfactory mixture prediction.
\end{abstract}


\section{Introduction}

Modeling olfactory mixtures remains one of the central challenges in machine olfaction. Although modern molecular representation learning methods such as the Principal Odor Map (POM)~\cite{Lee2023} provide strong embeddings for individual molecules, predicting perceptual similarity between mixtures requires reasoning over pairs of sets of odorants rather than over isolated molecules. A representation that is effective for single molecules is not automatically useful for mixture-level prediction, since the latter requires reasoning about relationships among the constituents of a mixture. Moreover, predicting mixture similarity requires reasoning about relationships between odorants across two mixtures.

Recent work such as POMMIX~\cite{Tom2025} showed that explicitly incorporating mixture structure within a deep learning framework can substantially improve predictive performance on olfactory mixture similarity tasks. By using mixture-aware architectures with attention-based interactions, POMMIX demonstrated that the task benefits from models that respect the internal structure of mixtures and the symmetry of mixture comparisons. We argue, however, that the main lesson of POMMIX is not simply that deep learning is superior, but that mixture-aware modeling is important.

This distinction is critical because classical baselines in previous work used relatively weak mixture representations. For example, a pair of mixtures can be represented as two independently aggregated descriptors that are then concatenated and passed to a regressor. Such representations summarize what each mixture is, but they do not directly encode how the two mixtures differ. As a result, they discard explicit cross-mixture comparison structure and force the downstream model to infer relationships indirectly from two separate descriptors, one for each mixture. In a low-data setting such as machine olfaction, this extra burden can degrade performance, and it can make classical baselines appear weaker than they might be under a more informative representation.

In this work, we revisit classical machine learning for olfactory mixture similarity in the low-data regime under substantially richer mixture representations. Our central thesis is that an important factor may not be deep versus classical learning per se, but whether the representation preserves explicit comparison structure between the two mixtures. More concretely, we ask whether a classical model becomes competitive with specialized deep learning models once it is given a feature map that directly exposes mixture differences and other comparison signals that respect the inherent symmetry in mixture similarity.

To test this hypothesis, we build symmetric feature representations from aggregated mixture descriptors using element-wise sums, products, and absolute differences, and we supplement them with a smaller block of cross-mixture molecular interaction statistics. We evaluate these representations using classical machine learning methods, primarily XGBoost and kernel ridge regression (KRR). For KRR, we compare a symmetric sample-product kernel with linear and RBF (radial basis function) kernels applied directly to the same explicit comparison vectors used by XGBoost. These experiments let us separate the role of model family from the role of representation quality.

Our results highlight the importance of representation in olfactory mixture learning. XGBoost with explicit difference-based symmetric features achieves performance competitive with POMMix, with a slightly higher mean Pearson correlation on the main benchmark. Ablation studies show that mixture-level comparison features alone achieve performance comparable to the combined representation, with no consistent additional improvement from the cross-mixture molecular interaction features. Feature-importance analysis further shows that coordinate-wise absolute differences are the strongest contributors to the XGBoost model. The sample-product KRR baseline remains weak, but KRR improves sharply when it is built on the explicit difference representation, although it still trails XGBoost. Overall, these results suggest that explicitly encoding how two mixtures differ can substantially improve classical models in low-data settings and may account for an important part of the advantage previously attributed to deep mixture-aware architectures.

\section{Related Work}

The literature most relevant to this work follows two connected lines of work: predicting odor perception from individual molecular structure, and extending such representations to multicomponent mixtures.

\subsection{From Single-Molecule Prediction to Learned Odor Representations}

Early quantitative structure--odor studies relied on physicochemical descriptors together with relatively simple statistical or machine-learning models. For example, Khan et al. showed that molecular physicochemical properties could be used to predict odor pleasantness \cite{Khan2007}. The DREAM Olfaction Prediction Challenge later provided a broader benchmark for predicting human perceptual attributes from molecular features. Keller et al. showed that machine-learning models based on chemoinformatic features could predict intensity, pleasantness, and several semantic odor descriptors \cite{Keller2017}. These studies established that molecular descriptors contain substantial perceptual signal. Subsequent work increasingly shifted toward learned molecular representations.

DeepNose learned low-dimensional features directly from three-dimensional molecular structure and used them to predict olfactory percepts \cite{Tran2019}. More recently, graph-based models have become particularly important because they can learn directly from molecular connectivity. The Principal Odor Map (POM) used a graph neural network to organize single molecules in a perceptually meaningful embedding space and demonstrated strong odor-quality prediction across several olfactory tasks \cite{Lee2023}. These learned embeddings provide useful representations for downstream prediction, but they do not by themselves determine how multiple molecules should be combined into a mixture representation or how two mixtures should be compared.

\subsection{Modeling Olfactory Mixtures}

Mixture modeling has developed along several complementary directions. Snitz et al. represented a multicomponent mixture by a single physicochemical structural vector and showed that distances between such vectors could predict perceptual similarity \cite{Snitz2013}. Ravia et al. subsequently refined this approach into a compact physicochemical measure for multicomponent odorants that predicts perceptual similarity and enabled the construction of olfactory metamers \cite{Ravia2020}. More recent work has moved toward learned mixture representations. Sisson et al. developed graph neural network models for predicting odor qualities of aroma-chemical blends \cite{Sisson2025}, while POMMIX combined POM molecular embeddings with attention-based mixture aggregation and a symmetric similarity head for mixture-similarity prediction \cite{Tom2025}.

Recent approaches further broaden the mixture-learning setting. AROMMA learns a unified embedding space for single molecules and two-molecule mixtures, using a chemical foundation model for molecular encoding and an attention-based aggregator for mixture composition \cite{Kang2026}. Most recently, Satarifard et al. reported results from the DREAM Olfactory Mixtures Prediction Challenge, which benchmarked perceptual-similarity prediction on a curated collection of odor-mixture pairs and combined top-performing approaches into a community model \cite{Satarifard2026}. Together, this literature shows a progression from fixed physicochemical mixture summaries toward learned, mixture-aware representations.

\subsection{Set and Pairwise Representations}

Olfactory mixture similarity has two distinct invariance requirements. First, each mixture is naturally represented as a set of molecules, so its representation should be invariant to the ordering of its constituents. This is an instance of the general problem of learning on sets. Prior work has developed both permutation-invariant representations of sets \cite{Zaheer2017} and kernels that operate directly on sets of vectors \cite{KondorJebara2003}.

Second, perceptual similarity is a symmetric relation between two mixtures: exchanging the two mixtures should not change the predicted similarity. More generally, learning relations between pairs of objects can exploit known structural properties of the relation. Waegeman et al.\ develop pairwise kernel methods that explicitly incorporate properties such as symmetry, including kernels designed for learning graded symmetric relations \cite{Waegeman2012}. These two levels of invariance, namely, within-mixture permutation invariance and between-mixture exchange symmetry, motivate representations that respect the structure of both the individual mixtures and the pair being compared.

\subsection{Classical Baselines and Pairwise Representations}

POMMIX also evaluated XGBoost as a classical baseline for olfactory mixture similarity prediction. In that baseline, molecular descriptors or pretrained molecular embeddings were aggregated separately within each mixture using Principal Neighbourhood Aggregation (PNA)-style statistics and then provided to XGBoost. Because the exchange symmetry of the two-mixture input was not explicitly encoded by the representation or model, the training data were augmented by permuting the mixture pairs \cite{Tom2025}. Despite using stronger molecular features such as frozen POM embeddings, this baseline remained substantially below CheMix and POMMIX.

Our work revisits this comparison at the representation level. Rather than describing the two mixtures only through separately aggregated descriptors, we explicitly encode symmetric pairwise comparison features, including coordinate-wise absolute differences, before fitting the classical learner. This lets us ask whether the performance gap between classical and mixture-aware deep models persists when the relationship between the two mixtures is represented more directly.

\section{Methods}

\subsection{Problem Setup}

Each training example is an unordered pair of olfactory mixtures, $M_A$ and $M_B$, together with a scalar perceptual dissimilarity label in $[0,1]$, where lower values indicate greater similarity. Because perceptual dissimilarity is symmetric, the prediction should be invariant with respect to exchanging $M_A$ and $M_B$. This makes symmetry an important design consideration for the representation, while still requiring the model to access informative differences between the mixtures.

\subsection{Molecular and Mixture Representations}

Each molecule is represented by a pretrained POM embedding. Because mixtures contain different numbers of molecules, we aggregate the molecular embeddings within each mixture into a fixed-length descriptor using PNA-style summary statistics. Specifically, for each embedding dimension, we compute the mean, variance, maximum, and minimum across the molecules in the mixture and concatenate these statistics into a single vector.

Let $f_A$ and $f_B$ denote the aggregated descriptors for mixtures $M_A$ and $M_B$, respectively. These descriptors summarize each mixture individually and serve as the inputs to the comparison features introduced below.

Although this aggregation produces a fixed-length representation suitable for standard machine learning models, it does not directly encode the relationship between the two mixtures. We therefore construct symmetric comparison features from $f_A$ and $f_B$ to explicitly represent how the mixtures differ.

\subsection{Mixture-Level Comparison Features}

Our first comparison representation, which we refer to as \emph{mixture-level comparison features}, is constructed directly from the aggregated mixture descriptors $f_A$ and $f_B$. It consists of three symmetric feature maps:
\[
\begin{aligned}
f_{\mathrm{sum}}     &= f_A + f_B,\\
f_{\mathrm{prod}}    &= f_A \odot f_B,\\
f_{\mathrm{absdiff}} &= \lvert f_A - f_B \rvert.
\end{aligned}
\]
where $\odot$ denotes element-wise multiplication.

These features are symmetric by construction: swapping $A$ and $B$ leaves all three unchanged. In particular, $f_{\mathrm{absdiff}}$ retains the coordinate-wise differences between the two aggregated mixture descriptors, making these differences directly available to the downstream model.

The full mixture-level comparison representation is obtained by concatenating the three components:

\begin{equation}
f_{\mathrm{mix}}
=
\left[
f_{\mathrm{sum}};
f_{\mathrm{prod}};
f_{\mathrm{absdiff}}
\right].
\end{equation}

\subsection{Cross-Mixture Molecular Interaction Features}

Our second feature block, which we refer to as \emph{cross-mixture molecular interaction features}, captures molecule-level interactions between the two mixtures. For each pair of molecules with one molecule from $M_A$ and the other from $M_B$, we compute the inner product and cosine similarity between their POM embeddings. We then summarize the resulting pairwise values using the minimum, maximum, mean, median, and population standard deviation. This gives ten global statistics: five for inner products and five for cosine similarities.

We also compute best-match statistics. Let $a_i$ and $b_j$ denote the molecular embeddings in $M_A$ and $M_B$, respectively, and define $S_{ij}=a_i^\top b_j$. For each molecule in one mixture, we identify its best match in the other mixture using the largest inner product. These statistics are computed separately from $M_A$ to $M_B$ and from $M_B$ to $M_A$:
\begin{align}
u &= \left[
\operatorname{mean}_i \max_j S_{ij},\;
\operatorname{std}_i \max_j S_{ij}
\right],\\
v &= \left[
\operatorname{mean}_j \max_i S_{ij},\;
\operatorname{std}_j \max_i S_{ij}
\right],
\end{align}
where $\operatorname{std}$ denotes population standard deviation.

The statistics computed over all cross-mixture molecule pairs are unchanged when $M_A$ and $M_B$ are exchanged. Exchanging the mixtures also interchanges $u$ and $v$. To make the best-match features symmetric, we therefore use
\begin{equation}
f_{\mathrm{best}}
=
\left[
u+v;\;
u\odot v;\;
|u-v|
\right].
\end{equation}
Since $u$ and $v$ each contain two components, this gives six symmetric best-match features. Together with the ten global statistics, they form a 16-dimensional cross-mixture molecular interaction representation. We use this symmetric interaction feature block to evaluate whether it provides additional information beyond the mixture-level comparison representation.

\begin{figure}[t]
    \centering
    \includegraphics[width=\linewidth]{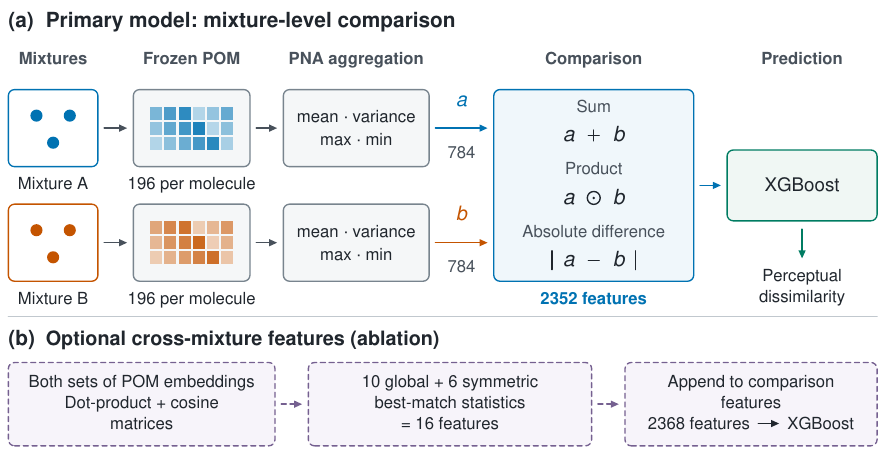}
    \caption{Overview of the comparison representation. (A) Frozen POM embeddings are aggregated separately within each mixture using the coordinate-wise mean, population variance, maximum, and minimum, excluding padding rows. The resulting 784-dimensional descriptors are compared using element-wise sum, product, and absolute difference, yielding 2352 features for the primary XGBoost model. Aggregation is invariant to molecule order, and the comparison representation is invariant to exchange of the two mixtures. (B) In the combined-model ablation, 16 symmetric cross-mixture molecular interaction features are appended to the comparison representation. Embedding glyphs are schematic rather than measured values.}
    \label{fig:comparison_method}
\end{figure}

Figure~\ref{fig:comparison_method} summarizes the primary comparison representation and the optional cross-mixture feature block.

\subsection{Models and Baselines}

We compare against two broad classes of existing baselines. The first consists of classical approaches based on relatively simple mixture representations without explicit interaction modeling. The Snitz baseline averages molecular descriptors within each mixture and computes the angle distance between the resulting mixture-level vectors. The XGBoost baselines use PNA-style aggregation of molecular descriptors to construct mixture-level representations before prediction. These preprocessing schemes construct representations for the two mixtures independently and therefore do not explicitly model cross-mixture interactions. Moreover, the symmetry of mixture similarity is not inherently encoded by XGBoost and is instead addressed through pair-swapping data augmentation. The second class consists of CheMix and POMMix, strong deep-learning baselines with interaction-aware mixture modeling. CheMix uses molecule-wise self-attention followed by permutation-invariant aggregation within each mixture and compares the resulting mixture embeddings using a scaled cosine prediction head. POMMix further fine-tunes the molecular POM encoder and CheMix model end-to-end.

Our approach constructs two complementary types of comparison features. Mixture-level comparison features explicitly capture symmetric relationships between aggregated mixture descriptors, whereas cross-mixture molecular interaction features capture molecule-level cross-mixture interactions. We evaluate these representations using XGBoost and additionally investigate kernel ridge regression (KRR) with symmetric kernels on unordered pairs of mixtures. Our goal is to evaluate whether explicitly modeling comparison and interaction structure can improve classical machine-learning models, relative both to classical baselines without such modeling and to deep-learning models with interaction-aware and symmetry-preserving inductive biases.

For kernel ridge regression, we use a symmetric sample-product kernel as a similarity-based baseline for unordered pairs of mixtures. This kernel compares two mixture pairs by combining the two possible mixture-to-mixture matchings, so it respects the fact that $(M_A,M_B)$ and $(M_B,M_A)$ represent the same input. However, it still compares samples mainly through mixture-level similarities and does not directly expose coordinate-wise differences within each mixture pair. We therefore also test explicit-difference KRR variants that apply linear or RBF kernels to the \texttt{absdiff}, mixture-level comparison, and combined comparison vectors. For these variants, the comparison vectors are standardized within each training fold before fitting KRR. The RBF kernel is defined as $k(u,v)=\exp(-\|u-v\|^2/\sigma^2)$. Hyperparameters are selected using validation Pearson correlation; the selected explicit-difference models use the RBF kernel.

Beyond this primary comparison, we conduct follow-up diagnostics using the same 2352-dimensional mixture-level comparison features and five data splits. We examine a Laplacian kernel, $k(u,v)=\exp(-\|u-v\|_1/\sigma)$, and a coordinate-wise quantile transformation to approximately standard normal marginals. Each preprocessing transformation is fitted using the training fold only, and KRR hyperparameters are selected by validation Pearson correlation. We also compare XGBoost and RBF KRR before and after an orthogonal rotation of the standardized features. These analyses investigate sensitivity to kernel choice, preprocessing, and feature orientation; they do not replace the primary model comparison. Full settings and results are provided in Appendix~\ref{app:krr_diagnostics}.

\subsection{Comparison with the Original POMMix Baselines}

We retain the released mixture-pair data, target values, and five-fold partitions used by POMMix \cite{Tom2025}. For the POM-based comparison, we use the same frozen molecular embeddings and PNA-style aggregation, concatenating the coordinate-wise mean, population variance, maximum, and minimum after excluding padding rows. The RDKit baseline instead uses the released RDKit descriptors.

The original XGBoost baseline aggregates each mixture independently and flattens the two descriptors in the ordering used by the released script. It uses maximum depth 1000 and training-only pair-swap augmentation. Our primary model uses explicit sum, product, and absolute-difference features, maximum depth 6, and no pair-swap augmentation because its representation is symmetric. Both use at most 1000 trees, learning rate 0.01, row and column subsampling rates of 0.8, and early stopping after 25 rounds without improvement in validation MSE. The released baseline script does not explicitly specify a training seed; our reproductions explicitly set it to 0.

We distinguish reproduction of the original baseline from a depth-matched control. The reproduction retains the original representation, augmentation, and depth limit. The control retains the representation and augmentation but uses depth 6. These comparisons examine whether depth explains the performance gap, but they do not fully isolate the effect of representation because symmetry and augmentation also differ. Implementation checks and provenance are given in Appendix~\ref{app:baseline_fidelity}.

\subsection{Evaluation Protocol}

We use the same released mixture-pair data and five-fold partitions as POMMix \cite{Tom2025}. The molecular POM representations used in the benchmark are pretrained on the filtered GoodScents--Leffingwell (GS-LF) dataset, which contains 4,814 molecules after preprocessing and is derived from an original collection of 4,983 molecules annotated with 138 odor descriptors \cite{Tom2025}. The mixture-similarity dataset is separate from GS-LF and contains 865 labeled mixture pairs compiled from three previous studies: 360 from Snitz, 260 from Bushdid, and 245 from Ravia \cite{Snitz2013,Bushdid2014,Ravia2020,Tom2025}.

For the primary benchmark, we use five random cross-validation splits stratified by study source, following the POMMix protocol \cite{Tom2025}. Each split contains 605 training examples, 87 validation examples, and 173 test examples, corresponding approximately to a 70\%/10\%/20\% train/validation/test split. XGBoost uses training seed 0 and early stopping based on validation MSE, whereas KRR hyperparameters are selected using validation Pearson correlation. Final performance is evaluated on the held-out test set. We report Pearson correlation ($\rho$) as the primary metric, together with RMSE and Kendall's $\tau$, and summarize performance as the mean $\pm$ sample standard deviation across the five splits ($\mathrm{ddof}=1$). No bootstrap resampling is used.

We verify that the explicit feature representations remain unchanged when we swap the two mixtures or reorder the molecules within either mixture. We test mixtures with different numbers of molecules, single-molecule mixtures, zero vectors, and padded inputs. Empty mixtures are rejected. Across all five test folds, swapping the mixtures produces a maximum absolute prediction difference of zero for both symmetric XGBoost and explicit-feature KRR, within the tolerance of $10^{-6}$. Neither model uses pair-swap augmentation.

To check how training randomness affects the results, we repeat the mixture-level and symmetric combined XGBoost experiments with training seeds 0--4. We keep the five data splits and all other training settings fixed, so this analysis isolates the effect of training randomness on the existing splits. We report results for every seed rather than selecting the seed with the best test performance.

\section{Results}

\subsection{Main Benchmark Comparison}

Table~\ref{tab:main_benchmark} summarizes the main random cross-validation benchmark, reporting mean performance and fold-level variation across five splits. The original classical baselines remain substantially below the strongest deep models, but this gap narrows sharply when XGBoost is paired with explicit comparison features. In particular, XGBoost using mixture-level comparison features achieves a mean Pearson correlation of 0.7861, compared with 0.7790 for POMMix, with overlapping fold-level variation. POMMix retains a higher mean Kendall correlation, so this comparison does not establish an advantage across all metrics.

Olfactory mixtures can exhibit non-additive perceptual effects, including masking and synergy, in which the perceptual contribution of one component depends on the presence and concentration of other components \cite{Ishii2008}. Related physiological studies have also observed nonlinear and antagonistic interactions among odorants at the level of olfactory sensory neurons \cite{Zak2020}. These observations motivate representations that retain information about relationships among mixture components rather than reducing each mixture to an independently aggregated descriptor. Although our benchmark does not directly identify specific masking or synergistic interactions, the strong performance of explicit comparison features is consistent with the broader view that relational structure is important for modeling mixture perception.

A second consideration is symmetry. Olfactory mixture similarity is invariant to the ordering of the two mixtures, and the representation of an individual mixture should likewise be invariant to the ordering of its constituent molecules \cite{Tom2025}. Deep mixture models such as CheMix encode these inductive biases explicitly through molecule-wise interaction modeling, permutation-invariant aggregation, and a symmetric cosine-based prediction head \cite{Tom2025}. Our results show that classical models can recover part of this advantage when relevant comparison structure is encoded explicitly in the input representation. In Table~\ref{tab:main_benchmark}, replacing RDKit descriptors with POM embeddings alone yields only a modest improvement for XGBoost, whereas introducing explicit symmetric comparison features produces a substantially larger gain. This suggests that the gap between the original classical baselines and specialized deep models reflects not only the quality of the molecular embeddings, but also how interaction and symmetry structure are incorporated into the representation.

Table~\ref{tab:matched_depth_baselines} compares the original POM input representation and the full mixture-level comparison representation at matched depth limits. At depth 6, their mean Pearson correlations are 0.5526 and 0.7861, respectively; at depth 1000, they are 0.5087 and 0.7831. Thus, changing the depth limit alone does not account for the performance gap. These comparisons use training seed 0; symmetry and pair-swap augmentation still differ between representations.

\begin{table}[t]
    \centering
    \caption{Main benchmark comparison using five stratified random cross-validation splits. Each split uses 70\%, 10\%, and 20\% of the samples for training, validation, and testing, respectively. The validation set is used for model and hyperparameter selection, and results are reported as mean $\pm$ sample standard deviation across the five test folds ($\mathrm{ddof}=1$). No bootstrap resampling is used.
    Explicit symmetric comparison features substantially strengthen classical learning and make XGBoost competitive with specialized deep mixture models.
    }
    \label{tab:main_benchmark}
    \begin{tabular}{lccc}
        \toprule
        Method & Pearson & RMSE & Kendall \\
        \midrule
        Snitz baseline & $0.3988 \pm 0.0568$ & $0.3344 \pm 0.0110$ & $0.2922 \pm 0.0471$ \\
        XGBoost RDKit & $0.4908 \pm 0.0585$ & $0.1642 \pm 0.0125$ & $0.3810 \pm 0.0442$ \\
        XGBoost POM embed & $0.5087 \pm 0.0480$ & $0.1636 \pm 0.0130$ & $0.4027 \pm 0.0433$ \\
        CheMix POM embed & $0.7465 \pm 0.0340$ & $0.1296 \pm 0.0075$ & $0.5449 \pm 0.0361$ \\
        POMMix & $0.7790 \pm 0.0314$ & $0.1184 \pm 0.0044$ & $\mathbf{0.5962 \pm 0.0251}$ \\
        XGBoost mixture-level & $\mathbf{0.7861 \pm 0.0359}$ & $\mathbf{0.1161 \pm 0.0070}$ & $0.5428 \pm 0.0399$ \\
        \bottomrule
    \end{tabular}
\end{table}

\begin{figure}[t]
    \centering
    \includegraphics[width=\linewidth]{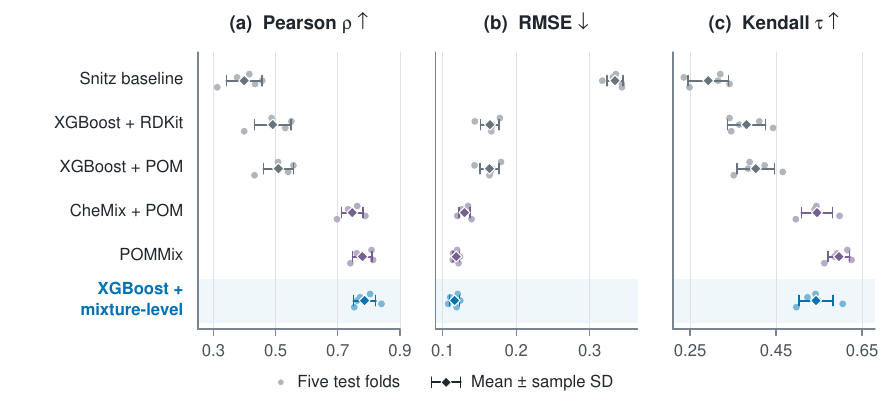}
    \caption{Main benchmark comparison across Pearson correlation, RMSE, and Kendall correlation. Small points denote individual test folds; diamonds and horizontal error bars indicate the five-fold mean and sample standard deviation ($\mathrm{ddof}=1$). Arrows indicate the favorable direction for each metric. Original XGBoost baselines are rerun with their original depth limit of 1000; the mixture-level model uses depth 6. All XGBoost results use training seed 0. Snitz, CheMix, and POMMix results are calculated from existing predictions. Differences in symmetry and augmentation remain between the original XGBoost baselines and our representation. These fold variations do not establish statistical equivalence or superiority.}
    \label{fig:benchmark_metrics}
\end{figure}

Figure~\ref{fig:benchmark_metrics} visualizes the fold-level results across the three evaluation metrics.

\begin{table}[t]
    \centering
    \small
    \caption{Depth-matched XGBoost comparisons using training seed 0. Values are mean $\pm$ sample standard deviation across the five test folds ($\mathrm{ddof}=1$). Independent POM retains the original pair representation and training-only pair-swap augmentation.}
    \label{tab:matched_depth_baselines}
    \begin{tabular}{llcc}
        \toprule
        Representation & Maximum depth & Pearson & RMSE \\
        \midrule
        Independent POM & 6 & $0.5526 \pm 0.0556$ & $0.1579 \pm 0.0133$ \\
        Full mixture-level & 6 & $0.7861 \pm 0.0359$ & $0.1161 \pm 0.0070$ \\
        Independent POM & 1000 & $0.5087 \pm 0.0480$ & $0.1636 \pm 0.0130$ \\
        Full mixture-level & 1000 & $0.7831 \pm 0.0365$ & $0.1170 \pm 0.0072$ \\
        \bottomrule
    \end{tabular}
\end{table}

\subsection{Ablation of Representation Components}

Table~\ref{tab:ablation} compares three XGBoost variants: mixture-level comparison features only, cross-mixture molecular interaction features only, and the combined model. Mixture-level comparison features alone achieve a mean Pearson correlation of 0.7861, compared with 0.7057 for cross-mixture molecular interaction features alone. Combining the two feature blocks gives a mean Pearson correlation of 0.7858, leaving performance nearly unchanged relative to mixture-level comparison features alone.

These ablations show that mixture-level comparison features capture nearly all of the performance of the combined representation. Adding the symmetric cross-mixture molecular interaction features does not improve mean test performance in the primary seed-0 experiments.

The follow-up analysis across five training seeds supports this interpretation. Averaging the five-fold Pearson means across seeds gives 0.78769 for mixture-level comparison features and 0.78787 for the combined representation. The combined model has a higher mean Pearson correlation for two seeds and a lower mean for three seeds. Thus, the small difference does not indicate a consistent improvement from adding the interaction features. Results for individual seeds are reported in Appendix~\ref{app:training_seeds}.

\begin{table}[t]
    \centering
    \caption{Ablation of representation components for XGBoost using training seed 0. Mixture-level and cross-mixture denote mixture-level comparison features and symmetric cross-mixture molecular interaction features, respectively; combined denotes their concatenation. Values are mean $\pm$ sample standard deviation across five folds ($\mathrm{ddof}=1$).}
    \label{tab:ablation}
    \small
    \begin{tabular}{lccc}
        \toprule
        Model & Pearson & RMSE & Validation Pearson \\
        \midrule
        Mixture-level only & $\mathbf{0.7861 \pm 0.0359}$ & $\mathbf{0.1161 \pm 0.0070}$ & $0.8142 \pm 0.0382$ \\
        Cross-mixture only & $0.7057 \pm 0.0255$ & $0.1329 \pm 0.0067$ & $0.7321 \pm 0.0459$ \\
        Combined & $0.7858 \pm 0.0362$ & $0.1162 \pm 0.0070$ & $\mathbf{0.8167 \pm 0.0414}$ \\
        \bottomrule
    \end{tabular}
\end{table}

\subsection{Ablation of Mixture-Level Comparison Features}

We next examine how much predictive performance is retained by the absolute-difference features alone. We compare three representations: 784 absolute-difference features, 1568 sum-and-product features, and the full 2352-dimensional mixture-level comparison representation. All three use XGBoost with maximum depth 6, training seed 0, and the same five data splits. The remaining settings are fixed at at most 1000 trees, learning rate 0.01, row and column subsampling rates of 0.8, and early stopping after 25 rounds without improvement in validation MSE. No cross-mixture molecular interaction features are included.

Table~\ref{tab:mixture_feature_ablation} shows that absolute differences alone achieve a mean Pearson correlation of 0.7709 and RMSE of 0.1195, compared with 0.7861 and 0.1161 for the full mixture-level representation. Relative to absolute differences alone, the full representation increases mean Pearson correlation by 0.0153 and reduces mean RMSE by 0.0034, with both metrics improving in all five folds. Using sums and products without absolute differences gives a lower mean Pearson correlation of 0.7395 and a higher mean RMSE of 0.1285.

These results show that absolute differences alone provide a strong representation under the current training protocol, while adding sums and products further improves performance. The ablation measures predictive performance; it does not establish that absolute differences account for a particular fraction of independent predictive signal or identify a biological mechanism.

\begin{table}[t]
    \centering
    \caption{Ablation of mixture-level comparison features using XGBoost with maximum depth 6 and training seed 0. No cross-mixture molecular interaction features are included. Values are mean $\pm$ sample standard deviation across the five test folds ($\mathrm{ddof}=1$).}
    \label{tab:mixture_feature_ablation}
    \begin{tabular}{lcc}
        \toprule
        Features & Pearson & RMSE \\
        \midrule
        Absolute differences only & $0.7709 \pm 0.0424$ & $0.1195 \pm 0.0071$ \\
        Sums and products only & $0.7395 \pm 0.0280$ & $0.1285 \pm 0.0070$ \\
        Full mixture-level & $\mathbf{0.7861 \pm 0.0359}$ & $\mathbf{0.1161 \pm 0.0070}$ \\
        \bottomrule
    \end{tabular}
\end{table}

\begin{figure}[t]
    \centering
    \includegraphics[width=\linewidth]{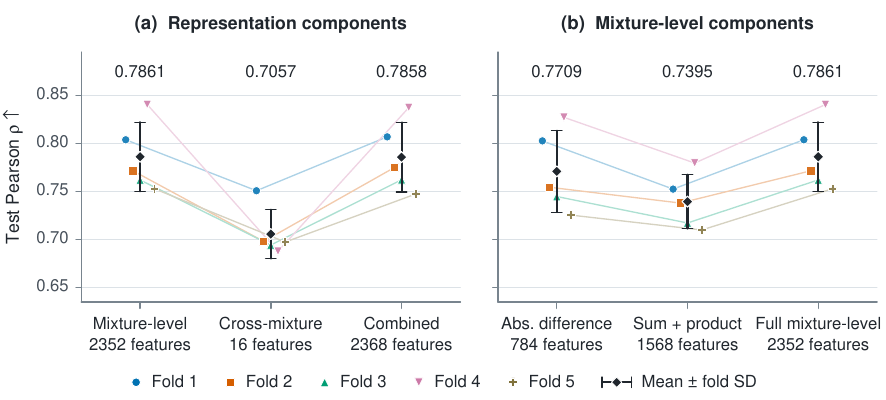}
    \caption{Ablation of explicit comparison features using depth-six XGBoost and training seed 0 on the same five folds. (A) Mixture-level comparison features, symmetric cross-mixture molecular interaction features, and their concatenation. (B) Absolute differences, sums and products, and the full mixture-level representation. Colored markers and thin lines identify corresponding folds; black diamonds and error bars denote the mean and sample standard deviation ($\mathrm{ddof}=1$). Numbers above each group are the mean Pearson correlations. Lines indicate paired evaluations rather than a sequential optimization procedure. All interaction features shown here are symmetric.}
    \label{fig:comparison_ablations}
\end{figure}

Figure~\ref{fig:comparison_ablations} shows the corresponding fold-paired ablation results.

\subsection{Sensitivity to Tree Depth}

To examine whether the advantage of absolute differences depends on restricting tree depth, we compare absolute differences alone, sums and products alone, and the full mixture-level representation at maximum depths ranging from 6 to 1000. Each configuration uses training seeds 0--4 on the original five folds, with all other training settings fixed. No cross-mixture molecular interaction features or pair-swap augmentation are used. For these summaries, metrics are averaged across seeds within each fold before calculating the mean across folds. The primary model remains at depth 6; this follow-up analysis is not used to select a new depth or training seed.

The mean Pearson gap between absolute differences and sums and products decreases from 0.03394 at depth 6 to 0.02892 at depth 12, but increases slightly to 0.03019 at depth 32. Thus, the gap does not decrease monotonically. For sums and products, increasing the depth limit from 6 to 1000 reduces mean training RMSE from 0.015325 to 0.009111, while mean test RMSE changes from 0.128635 to 0.129236. Depth limits of 50, 100, 300, and 1000 produce identical test predictions; the deepest selected tree reaches depth 34. These results indicate that the advantage of explicit absolute differences cannot be explained solely by the depth-six restriction under the current training protocol. Complete curves, fold-level variation, and training diagnostics are reported in Appendices~\ref{app:depth_sensitivity} and~\ref{app:dense_depth}.

\subsection{Feature Importance Reveals a Dominant Comparison Signal}

Given the strong performance of mixture-level comparison features in the ablation study, we next examine which components of the mixture-level comparison representation contribute most strongly to the fitted XGBoost models. Table~\ref{tab:feature_importance_family} shows that the absolute-difference family accounts for the largest share of normalized gain importance in the symmetric combined model. This pattern is consistent across folds: absolute-difference features account for roughly 50\% to 56\% of the total importance, with a mean share of 0.5243. Product features contribute about 23\% to 27\% (mean 0.2487), while sum features contribute about 20\% to 24\% (mean 0.2159). In contrast, all cross-mixture molecular interaction features together account for approximately 1.1\% on average (mean 0.0111).

The individual feature rankings show the same pattern at a finer level. Across folds, 16 to 20 of the 20 highest-ranked features are from the \texttt{sym\_absdiff\_*} family, with the remaining features drawn from the \texttt{sym\_prod\_*} and \texttt{cross\_*} families. Thus, the importance of the mixture-level comparison representation is not distributed evenly across its components: coordinate-wise absolute differences account for the largest share of gain importance.

One possible explanation is that absolute differences directly measure how much the two mixtures differ in each feature dimension. These features may therefore make it easier for XGBoost to learn the relationship between mixture differences and similarity ratings. Sum and product features describe the pair but do not express these differences directly. Together with the results in Table~\ref{tab:mixture_feature_ablation}, this analysis suggests that explicit differences between the aggregated mixture descriptors are useful for prediction.

\begin{table}[t]
    \centering
    \caption{Normalized gain importance by feature family in the symmetric combined XGBoost model using training seed 0. Values are mean $\pm$ sample standard deviation across five folds ($\mathrm{ddof}=1$).}
    \label{tab:feature_importance_family}
    \begin{tabular}{lc}
        \toprule
        Feature family & Gain share \\
        \midrule
        \texttt{sym\_absdiff\_*} & $\mathbf{0.5243 \pm 0.0280}$ \\
        \texttt{sym\_prod\_*} & $0.2487 \pm 0.0159$ \\
        \texttt{sym\_sum\_*} & $0.2159 \pm 0.0141$ \\
        \texttt{cross\_*} & $0.0111 \pm 0.0086$ \\
        \bottomrule
    \end{tabular}
\end{table}


\subsection{Explicit Comparison Features Improve Kernel Models}

We next test whether KRR benefits from the same explicit comparison structure that improves XGBoost. Table~\ref{tab:kernel_comparison} compares the similarity-based sample-product kernel with KRR models applied directly to explicit comparison features. The sample-product kernel achieves a mean Pearson correlation of $0.4521 \pm 0.0226$, substantially below the tree-based models.

Using explicit comparison features produces a large improvement. KRR with the full mixture-level comparison representation reaches a mean Pearson correlation of $0.7504 \pm 0.0343$, while the \texttt{absdiff}-only representation reaches $0.7403 \pm 0.0581$. Both substantially outperform the sample-product kernel. Adding symmetric cross-mixture molecular interaction features to mixture-level comparison features does not provide an additional gain, with the combined representation reaching $0.7478 \pm 0.0375$.

These results show that the benefit of explicit comparison features is not limited to XGBoost. KRR also improves substantially when the relationship between the two mixtures is represented directly. However, the best KRR model in the primary comparison remains below XGBoost, motivating further examination of kernel choice and preprocessing. 

In follow-up diagnostics on the same five splits, Laplacian KRR achieves a mean Pearson correlation of $0.7673 \pm 0.0423$. Combining this kernel with a training-fitted quantile transformation increases the mean correlation to $0.7789 \pm 0.0370$, compared with $0.7504 \pm 0.0343$ for the original RBF KRR. The improvement over the original RBF model occurs in all five folds. A separate XGBoost control trained on standardized inputs achieves $0.7866 \pm 0.0370$, leaving a mean Pearson difference of $0.0077$. This control is distinct from the original XGBoost result in Table~\ref{tab:kernel_comparison}. The smaller difference does not establish statistical equivalence.

A rotation diagnostic further indicates that feature orientation matters for XGBoost: a fixed orthogonal rotation of the standardized inputs reduces its mean Pearson correlation from $0.7866$ to $0.7561$. RBF KRR remains unchanged under the same rotation when compared using the corresponding bandwidth settings, as expected from preservation of Euclidean distances. These findings suggest that kernel geometry and preprocessing contribute to the observed performance gap, but do not isolate smoothness as its cause. Related tabular-data studies also emphasize feature orientation and feature distributions when comparing trees with neural networks \cite{Grinsztajn2022,McElfresh2023}; they provide context rather than direct evidence for the mechanism in our KRR experiments. These exploratory diagnostics are detailed in Appendix~\ref{app:krr_diagnostics}.

\begin{table}[t]
    \centering
    \caption{Similarity-based and explicit-difference KRR compared with tree-based models. Values are mean $\pm$ sample standard deviation over five random CV splits ($\mathrm{ddof}=1$). KRR improves substantially once it is built on explicit comparison representations, although XGBoost remains strongest. Mixture-level denotes mixture-level comparison features; combined additionally includes symmetric cross-mixture molecular interaction features. XGBoost results use training seed 0.}
    \label{tab:kernel_comparison}
    \small
    \begin{tabular}{lccc}
        \toprule
        Model & Pearson & RMSE & Validation Pearson \\
        \midrule
        KRR sample-product kernel & $0.4521 \pm 0.0226$ & $0.2094 \pm 0.0569$ & $0.5815 \pm 0.0625$ \\
        KRR explicit \texttt{absdiff} kernel & $0.7403 \pm 0.0581$ & $0.1262 \pm 0.0075$ & $0.7889 \pm 0.0445$ \\
        KRR explicit mixture-level kernel & $0.7504 \pm 0.0343$ & $0.1245 \pm 0.0059$ & $0.7872 \pm 0.0574$ \\
        KRR explicit combined kernel & $0.7478 \pm 0.0375$ & $0.1259 \pm 0.0086$ & $0.7869 \pm 0.0582$ \\
        XGBoost mixture-level only & $\mathbf{0.7861 \pm 0.0359}$ & $\mathbf{0.1161 \pm 0.0070}$ & $0.8142 \pm 0.0382$ \\
        XGBoost combined & $0.7858 \pm 0.0362$ & $0.1162 \pm 0.0070$ & $\mathbf{0.8167 \pm 0.0414}$ \\
        \bottomrule
    \end{tabular}
\end{table}


\section{Discussion and Future Directions}

Our results refine rather than contradict the main conclusion of POMMix. POMMix showed that incorporating mixture structure and task symmetry can substantially improve olfactory mixture similarity prediction. Our experiments support the importance of this structure, while suggesting that part of the gain can be obtained without a specialized deep mixture architecture when the relationship between the two mixtures is represented explicitly.

This distinction is visible in the classical baselines. Replacing RDKit descriptors with POM embeddings alone produces only a modest improvement for the original XGBoost baseline. In contrast, constructing explicit symmetric comparison features from the aggregated POM representations makes XGBoost competitive with POMMix on the main cross-validation benchmark. The ablation and feature-importance results further localize this improvement: most of the gain is captured by mixture-level comparison features, and coordinate-wise absolute differences account for the largest share of feature importance.

The kernel experiments provide a complementary view. A symmetric sample-product kernel respects the exchange symmetry of the mixture pair but performs poorly when the pair is represented only through mixture-to-mixture similarity values. Applying KRR to the explicit mixture-level comparison representation substantially improves performance. This suggests that satisfying the symmetry of the task is useful but not sufficient; the form in which the comparison information is exposed to the learner also matters. The follow-up diagnostics further suggest that the remaining gap depends on kernel choice and preprocessing, rather than representing a general limitation of KRR.

Taken together, these results support a representation-centered interpretation of the benchmark. In this low-data setting, explicitly encoding the relationship between the two mixtures can substantially reduce the gap between classical and specialized deep models. This does not imply that model architecture is unimportant, particularly under other generalization settings, but it shows that model-family comparisons can depend strongly on the representation supplied to each learner.

Given that the difference in mean Pearson correlation between the best XGBoost model and POMMix is small relative to the variation across folds, we interpret the result as evidence that XGBoost is competitive with POMMix under the current representation and evaluation setting, rather than as evidence of a general performance advantage.

An important direction for future work is to evaluate whether the advantage of explicit comparison features persists beyond random cross-validation. In particular, our classical models could be evaluated under more challenging generalization settings involving unseen molecules and shifts in mixture size, analogous to the leave-molecules-out and mixture-size splits considered by POMMix \cite{Tom2025}. These experiments would help determine whether the gains from explicit comparison features reflect robust inductive biases or are specific to the current random-split benchmark. Additional downstream evaluations of the learned representations could further test whether their utility extends beyond pairwise similarity prediction.

A second direction is to develop richer representations of molecule-level interactions. The relatively small contribution of cross-mixture molecular interaction features in the present experiments may reflect the limited summary statistics used to compress cross-mixture molecular interactions rather than an absence of interaction effects. Future work could investigate more expressive but still symmetry-preserving interaction representations that retain finer-grained information about individual molecule pairs and capture nonlinear or non-additive effects that may be lost during aggregation.

Finally, the kernel results motivate further study of learning methods designed specifically for structured pairwise inputs. Our experiments consider only a limited set of symmetric KRR formulations, and the substantial improvement obtained by applying KRR to mixture-level comparison features suggests that kernel performance is also sensitive to how comparison structure is represented. More expressive permutation-invariant kernels, or kernels that operate directly on sets of molecular embeddings and their cross-mixture interactions, may provide a useful middle ground between hand-constructed comparison features and end-to-end deep mixture models.

\section{Conclusion}

We revisited classical learning for olfactory mixture similarity prediction by separating the choice of model from the representation of the mixture pair. Explicit symmetric comparison features substantially improve XGBoost, allowing it to match the performance of specialized deep mixture models on the main five-fold cross-validation benchmark. The same representation also produces a large improvement in KRR, while ablation and feature-importance analyses identify coordinate-wise mixture differences as the strongest component of the current representation.

These results suggest that explicit pairwise representation is an important factor in low-data olfactory mixture learning. Rather than viewing the performance of mixture-aware models primarily as a consequence of model complexity, our findings highlight the importance of how the relationship between two mixtures is encoded. Future work should test whether this conclusion extends to more challenging generalization settings and explore comparison-aware kernels and other models that can make better use of structured mixture representations.

\clearpage
\appendix
\input{appendix}

\bibliographystyle{unsrt}
\bibliography{references}

\end{document}

%% file: appendix.tex
\section{Sensitivity to Training Seeds}
\label{app:training_seeds}

We repeated the mixture-level and symmetric combined XGBoost experiments using training seeds 0--4 while retaining the same five data splits. All other settings were fixed: maximum tree depth 6, at most 1000 trees, learning rate 0.01, row and column subsampling rates of 0.8, and early stopping after 25 rounds without improvement in validation MSE. Neither representation used pair-swap augmentation.

Table~\ref{tab:training_seeds} reports the five-fold mean and sample standard deviation for each seed. The relative ranking changes across seeds. Averaging the five-fold means across seeds gives Pearson correlations of 0.78769 and 0.78787 for the mixture-level and combined representations, respectively. The corresponding mean RMSE values are 0.115795 and 0.115745. These small differences support comparable performance rather than a consistent additional improvement from the interaction features.

This follow-up analysis evaluates training randomness on fixed data splits. It does not assess variability under new data partitions, and the 25 seed--fold evaluations should not be treated as independent datasets. No seed was selected based on test performance.

\begin{table}[H]
    \centering
    \small
    \caption{Sensitivity to training seeds on the fixed five-fold splits. Each entry is the mean $\pm$ sample standard deviation across five folds for one seed ($\mathrm{ddof}=1$). Mixture-level denotes mixture-level comparison features; combined additionally includes symmetric cross-mixture molecular interaction features.}
    \label{tab:training_seeds}
    \begin{tabular}{lcc}
        \toprule
        Training seed & Mixture-level & Combined \\
        \midrule
        \multicolumn{3}{c}{Pearson correlation} \\
        \midrule
        0 & $0.78615 \pm 0.03591$ & $0.78576 \pm 0.03617$ \\
        1 & $0.78655 \pm 0.03521$ & $0.78933 \pm 0.03288$ \\
        2 & $0.78764 \pm 0.03775$ & $0.78770 \pm 0.03707$ \\
        3 & $0.78848 \pm 0.03514$ & $0.78776 \pm 0.03560$ \\
        4 & $0.78961 \pm 0.03686$ & $0.78880 \pm 0.03830$ \\
        \midrule
        \multicolumn{3}{c}{RMSE} \\
        \midrule
        0 & $0.116127 \pm 0.007029$ & $0.116194 \pm 0.006978$ \\
        1 & $0.116028 \pm 0.007399$ & $0.115457 \pm 0.006873$ \\
        2 & $0.115751 \pm 0.007860$ & $0.115827 \pm 0.007680$ \\
        3 & $0.115724 \pm 0.006700$ & $0.115765 \pm 0.007496$ \\
        4 & $0.115344 \pm 0.007584$ & $0.115482 \pm 0.008033$ \\
        \bottomrule
    \end{tabular}
\end{table}

\section{Sensitivity to Tree Depth}
\label{app:depth_sensitivity}

We evaluated maximum depths 6, 8, 10, 20, 50, 100, 300, and 1000 using five training seeds, five fixed folds, and three feature representations, giving 600 configurations. After completing depths 6, 8, and 10, the remaining intermediate depths were revised to cover a wider range, including 100 and 300. All completed results at depths 6, 8, and 10 were retained. The additional dense-grid experiments at depths 12, 16, 24, and 32 are reported in Appendix~\ref{app:dense_depth}. No depth was removed because of its test score.

Existing models were reused only after checking dependencies, parameters, feature names, split indices, and predictions. The primary depth-six model was retained; neither the depth nor the seed was selected using test performance. Every configuration used at most 1000 trees, learning rate 0.01, row and column subsampling rates of 0.8, and early stopping after 25 rounds without improvement in validation MSE. No cross-mixture molecular interaction features or pair-swap augmentation were used. The labels and data partitions were unchanged.

For the overall summaries, metrics are averaged across seeds within each fold before calculating the mean and sample standard deviation across five folds ($\mathrm{ddof}=1$). Individual-seed tables instead report the mean and sample standard deviation across that seed's five folds. The 25 seed--fold evaluations are not independent datasets. No averaging of predictions across seeds is performed.

\begin{figure}[H]
    \centering
    \includegraphics[width=\linewidth]{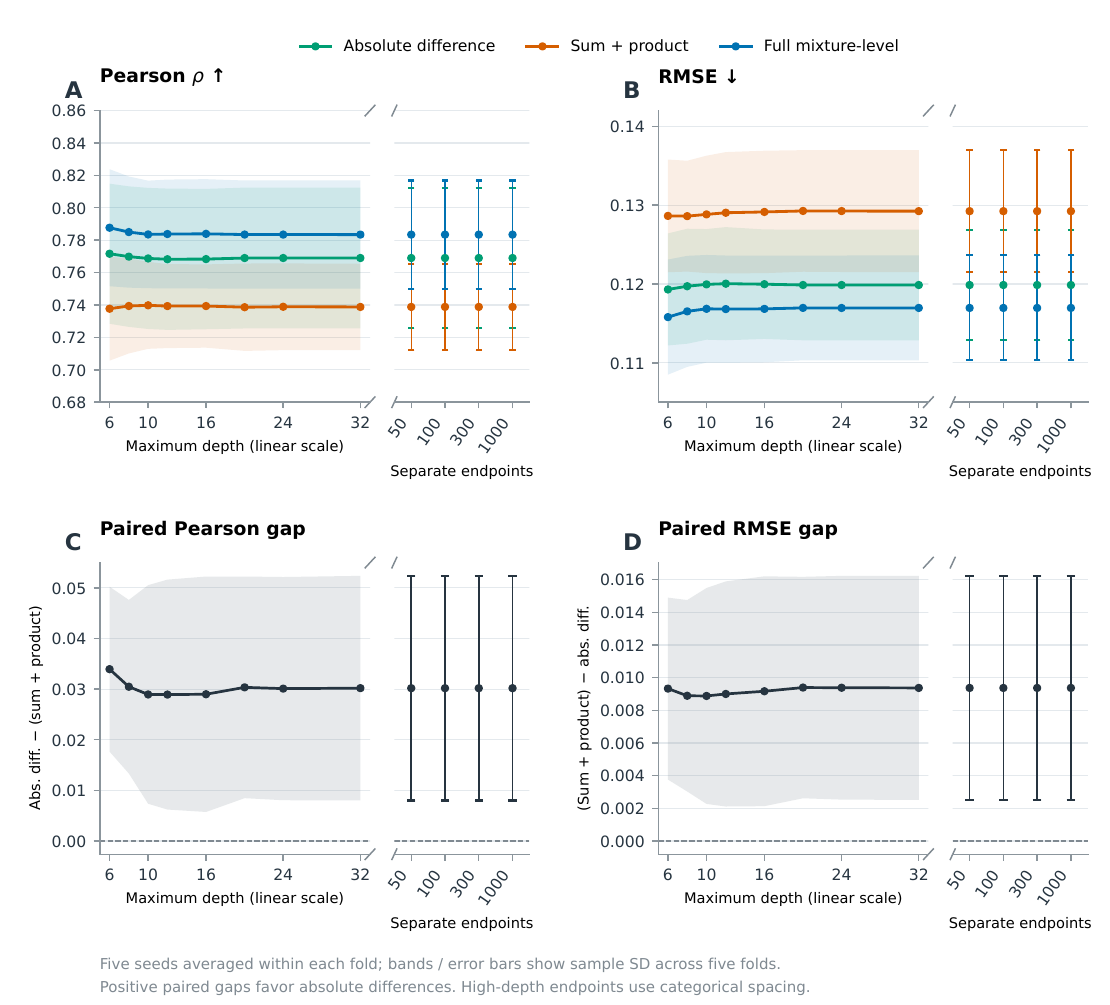}
    \caption{Tree-depth sensitivity for absolute differences, sums and products, and the full mixture-level representation. All 12 evaluated maximum depths are retained: 6, 8, 10, 12, 16, 20, 24, 32, 50, 100, 300, and 1000. (A,B) Test Pearson correlation and RMSE. (C,D) Fold-paired gaps, oriented so positive values favor absolute differences over sums and products. At each depth, metrics are first averaged across training seeds 0--4 within each fold, and then summarized across the five folds. Bands and error bars denote sample standard deviations across folds ($\mathrm{ddof}=1$); the 25 seed--fold evaluations are not treated as independent datasets. Depths 6--32 use a linear numerical axis; 50, 100, 300, and 1000 are displayed as separately spaced endpoints beyond an axis break. The primary model remains at depth 6, and these diagnostics are not used to select a new main model.}
    \label{fig:depth_sensitivity_all}
\end{figure}

Figure~\ref{fig:depth_sensitivity_all} summarizes performance and paired feature gaps across all evaluated depth limits.

\begin{table}[H]
    \centering
    \small
    \setlength{\tabcolsep}{4pt}
    \caption{Test performance after averaging metrics over training seeds within each fold. Values are mean $\pm$ sample standard deviation across five folds.}
    \label{tab:depth_overall}
    \begin{tabular}{llcc}
        \toprule
        Depth & Features & Pearson & RMSE \\
        \midrule
        6 & Absolute differences & $0.77163 \pm 0.04323$ & $0.11931 \pm 0.00711$ \\
        6 & Sums and products & $0.73769 \pm 0.03205$ & $0.12863 \pm 0.00715$ \\
        6 & Full mixture-level & $0.78769 \pm 0.03610$ & $0.11579 \pm 0.00730$ \\
        8 & Absolute differences & $0.76984 \pm 0.04337$ & $0.11971 \pm 0.00730$ \\
        8 & Sums and products & $0.73937 \pm 0.02927$ & $0.12860 \pm 0.00703$ \\
        8 & Full mixture-level & $0.78500 \pm 0.03431$ & $0.11652 \pm 0.00704$ \\
        10 & Absolute differences & $0.76873 \pm 0.04354$ & $0.11995 \pm 0.00702$ \\
        10 & Sums and products & $0.73978 \pm 0.02685$ & $0.12883 \pm 0.00745$ \\
        10 & Full mixture-level & $0.78355 \pm 0.03319$ & $0.11685 \pm 0.00683$ \\
        20 & Absolute differences & $0.76897 \pm 0.04347$ & $0.11987 \pm 0.00702$ \\
        20 & Sums and products & $0.73863 \pm 0.02696$ & $0.12926 \pm 0.00771$ \\
        20 & Full mixture-level & $0.78347 \pm 0.03337$ & $0.11696 \pm 0.00665$ \\
        50 & Absolute differences & $0.76898 \pm 0.04344$ & $0.11987 \pm 0.00702$ \\
        50 & Sums and products & $0.73880 \pm 0.02665$ & $0.12924 \pm 0.00774$ \\
        50 & Full mixture-level & $0.78344 \pm 0.03341$ & $0.11696 \pm 0.00666$ \\
        100 & Absolute differences & $0.76898 \pm 0.04344$ & $0.11987 \pm 0.00702$ \\
        100 & Sums and products & $0.73880 \pm 0.02665$ & $0.12924 \pm 0.00774$ \\
        100 & Full mixture-level & $0.78344 \pm 0.03341$ & $0.11696 \pm 0.00666$ \\
        300 & Absolute differences & $0.76898 \pm 0.04344$ & $0.11987 \pm 0.00702$ \\
        300 & Sums and products & $0.73880 \pm 0.02665$ & $0.12924 \pm 0.00774$ \\
        300 & Full mixture-level & $0.78344 \pm 0.03341$ & $0.11696 \pm 0.00666$ \\
        1000 & Absolute differences & $0.76898 \pm 0.04344$ & $0.11987 \pm 0.00702$ \\
        1000 & Sums and products & $0.73880 \pm 0.02665$ & $0.12924 \pm 0.00774$ \\
        1000 & Full mixture-level & $0.78344 \pm 0.03341$ & $0.11696 \pm 0.00666$ \\
        \bottomrule
    \end{tabular}
\end{table}



\begin{table}[H]
    \centering
    \small
    \setlength{\tabcolsep}{4pt}
    \caption{Paired gaps: Pearson is absdiff minus sum+product; RMSE is sum+product minus absdiff. Positive gaps favor absdiff. The last column is the mean Pearson gap minus its depth-six value. Statistics use within-fold seed averaging followed by fold mean and sample standard deviation.}
    \label{tab:depth_gaps}
    \begin{tabular}{lccc}
        \toprule
        Depth & Pearson gap & RMSE gap & Gap change \\
        \midrule
        6 & $0.03394 \pm 0.01628$ & $0.009329 \pm 0.005569$ & $+0.00000$ \\
        8 & $0.03047 \pm 0.01714$ & $0.008890 \pm 0.005861$ & $-0.00347$ \\
        10 & $0.02894 \pm 0.02159$ & $0.008879 \pm 0.006609$ & $-0.00500$ \\
        20 & $0.03035 \pm 0.02189$ & $0.009393 \pm 0.006779$ & $-0.00359$ \\
        50 & $0.03018 \pm 0.02219$ & $0.009370 \pm 0.006857$ & $-0.00376$ \\
        100 & $0.03018 \pm 0.02219$ & $0.009370 \pm 0.006857$ & $-0.00376$ \\
        300 & $0.03018 \pm 0.02219$ & $0.009370 \pm 0.006857$ & $-0.00376$ \\
        1000 & $0.03018 \pm 0.02219$ & $0.009370 \pm 0.006857$ & $-0.00376$ \\
        \bottomrule
    \end{tabular}
\end{table}

\begin{table}[H]
    \centering
    \small
    \setlength{\tabcolsep}{4pt}
    \caption{Pearson gap for each fixed fold after averaging paired gaps across seeds. Positive values favor absdiff.}
    \label{tab:depth_fold_pearson_gap}
    \begin{tabular}{lccccc}
        \toprule
        Depth & cv0 & cv1 & cv2 & cv3 & cv4 \\
        \midrule
        6 & $+0.045112$ & $+0.010859$ & $+0.034788$ & $+0.052528$ & $+0.026406$ \\
        8 & $+0.040733$ & $+0.012702$ & $+0.030148$ & $+0.053408$ & $+0.015365$ \\
        10 & $+0.041745$ & $+0.003703$ & $+0.026984$ & $+0.057997$ & $+0.014277$ \\
        20 & $+0.045886$ & $+0.003191$ & $+0.025647$ & $+0.058284$ & $+0.018730$ \\
        50 & $+0.046062$ & $+0.002838$ & $+0.024656$ & $+0.058605$ & $+0.018733$ \\
        100 & $+0.046062$ & $+0.002838$ & $+0.024656$ & $+0.058605$ & $+0.018733$ \\
        300 & $+0.046062$ & $+0.002838$ & $+0.024656$ & $+0.058605$ & $+0.018733$ \\
        1000 & $+0.046062$ & $+0.002838$ & $+0.024656$ & $+0.058605$ & $+0.018733$ \\
        \bottomrule
    \end{tabular}
\end{table}

\begin{table}[H]
    \centering
    \small
    \setlength{\tabcolsep}{4pt}
    \caption{RMSE gap for each fixed fold after averaging paired gaps across seeds. Positive values favor absdiff.}
    \label{tab:depth_fold_rmse_gap}
    \begin{tabular}{lccccc}
        \toprule
        Depth & cv0 & cv1 & cv2 & cv3 & cv4 \\
        \midrule
        6 & $+0.013110$ & $+0.003359$ & $+0.007501$ & $+0.016937$ & $+0.005740$ \\
        8 & $+0.012397$ & $+0.003873$ & $+0.006955$ & $+0.017331$ & $+0.003894$ \\
        10 & $+0.013074$ & $+0.002292$ & $+0.006845$ & $+0.018156$ & $+0.004028$ \\
        20 & $+0.014160$ & $+0.002231$ & $+0.007083$ & $+0.018579$ & $+0.004914$ \\
        50 & $+0.014242$ & $+0.002179$ & $+0.006956$ & $+0.018651$ & $+0.004823$ \\
        100 & $+0.014242$ & $+0.002179$ & $+0.006956$ & $+0.018651$ & $+0.004823$ \\
        300 & $+0.014242$ & $+0.002179$ & $+0.006956$ & $+0.018651$ & $+0.004823$ \\
        1000 & $+0.014242$ & $+0.002179$ & $+0.006956$ & $+0.018651$ & $+0.004823$ \\
        \bottomrule
    \end{tabular}
\end{table}

\begin{table}[H]
    \centering
    \small
    \setlength{\tabcolsep}{4pt}
    \caption{Training diagnostics. Training RMSE and selected-tree counts use within-fold seed averaging and report mean $\pm$ fold sample standard deviation. Actual depth gives the range of the deepest selected tree across all 25 seed--fold configurations.}
    \label{tab:depth_diagnostics}
    \begin{tabular}{llccc}
        \toprule
        Limit & Features & Train RMSE & Actual depth & Selected trees \\
        \midrule
        6 & Absolute differences & $0.03086 \pm 0.00697$ & 6--6 & $552.6 \pm 216.0$ \\
        6 & Sums and products & $0.01533 \pm 0.00359$ & 6--6 & $756.8 \pm 152.1$ \\
        6 & Full mixture-level & $0.01716 \pm 0.00483$ & 6--6 & $578.6 \pm 155.9$ \\
        8 & Absolute differences & $0.02737 \pm 0.00500$ & 8--8 & $578.6 \pm 250.4$ \\
        8 & Sums and products & $0.01030 \pm 0.00267$ & 8--8 & $767.1 \pm 140.1$ \\
        8 & Full mixture-level & $0.01220 \pm 0.00424$ & 8--8 & $613.8 \pm 190.7$ \\
        10 & Absolute differences & $0.02698 \pm 0.00546$ & 10--10 & $567.2 \pm 239.3$ \\
        10 & Sums and products & $0.00959 \pm 0.00285$ & 10--10 & $728.3 \pm 154.5$ \\
        10 & Full mixture-level & $0.01068 \pm 0.00339$ & 10--10 & $603.4 \pm 128.8$ \\
        20 & Absolute differences & $0.02665 \pm 0.00545$ & 18--20 & $558.8 \pm 219.6$ \\
        20 & Sums and products & $0.00917 \pm 0.00306$ & 20--20 & $684.7 \pm 122.6$ \\
        20 & Full mixture-level & $0.01110 \pm 0.00404$ & 18--20 & $552.5 \pm 131.7$ \\
        50 & Absolute differences & $0.02665 \pm 0.00545$ & 18--28 & $552.8 \pm 216.3$ \\
        50 & Sums and products & $0.00911 \pm 0.00307$ & 24--34 & $685.2 \pm 120.6$ \\
        50 & Full mixture-level & $0.01101 \pm 0.00390$ & 18--28 & $554.6 \pm 128.5$ \\
        100 & Absolute differences & $0.02665 \pm 0.00545$ & 18--28 & $552.8 \pm 216.3$ \\
        100 & Sums and products & $0.00911 \pm 0.00307$ & 24--34 & $685.2 \pm 120.6$ \\
        100 & Full mixture-level & $0.01101 \pm 0.00390$ & 18--28 & $554.6 \pm 128.5$ \\
        300 & Absolute differences & $0.02665 \pm 0.00545$ & 18--28 & $552.8 \pm 216.3$ \\
        300 & Sums and products & $0.00911 \pm 0.00307$ & 24--34 & $685.2 \pm 120.6$ \\
        300 & Full mixture-level & $0.01101 \pm 0.00390$ & 18--28 & $554.6 \pm 128.5$ \\
        1000 & Absolute differences & $0.02665 \pm 0.00545$ & 18--28 & $552.8 \pm 216.3$ \\
        1000 & Sums and products & $0.00911 \pm 0.00307$ & 24--34 & $685.2 \pm 120.6$ \\
        1000 & Full mixture-level & $0.01101 \pm 0.00390$ & 18--28 & $554.6 \pm 128.5$ \\
        \bottomrule
    \end{tabular}
\end{table}

For every matched seed--fold--representation configuration, depth limits 50, 100, 300, and 1000 give identical saved test predictions. Their individual-seed results are therefore reported together in Table~\ref{tab:depth_seed_plateau}. The raw per-configuration predictions, paired depth differences, learning curves, and model files are retained with the reproducibility artifacts.

\begin{table}[H]
    \centering
    \small
    \setlength{\tabcolsep}{4pt}
    \caption{Individual training seeds at maximum depth 6. Each entry is the mean $\pm$ sample standard deviation across the five fixed test folds.}
    \label{tab:depth_seed_6}
    \begin{tabular}{llcc}
        \toprule
        Seed & Features & Pearson & RMSE \\
        \midrule
        0 & Absolute differences & $0.77089 \pm 0.04240$ & $0.119489 \pm 0.007080$ \\
        0 & Sums and products & $0.73947 \pm 0.02803$ & $0.128455 \pm 0.006974$ \\
        0 & Full mixture-level & $0.78615 \pm 0.03591$ & $0.116127 \pm 0.007029$ \\
        1 & Absolute differences & $0.77388 \pm 0.04180$ & $0.118834 \pm 0.007125$ \\
        1 & Sums and products & $0.73783 \pm 0.03414$ & $0.128575 \pm 0.008131$ \\
        1 & Full mixture-level & $0.78655 \pm 0.03521$ & $0.116028 \pm 0.007399$ \\
        2 & Absolute differences & $0.76961 \pm 0.04417$ & $0.119659 \pm 0.007075$ \\
        2 & Sums and products & $0.73730 \pm 0.03213$ & $0.128747 \pm 0.006728$ \\
        2 & Full mixture-level & $0.78764 \pm 0.03775$ & $0.115751 \pm 0.007860$ \\
        3 & Absolute differences & $0.77002 \pm 0.04438$ & $0.119559 \pm 0.007121$ \\
        3 & Sums and products & $0.73801 \pm 0.03548$ & $0.128622 \pm 0.007420$ \\
        3 & Full mixture-level & $0.78848 \pm 0.03514$ & $0.115724 \pm 0.006700$ \\
        4 & Absolute differences & $0.77376 \pm 0.04381$ & $0.118985 \pm 0.007298$ \\
        4 & Sums and products & $0.73586 \pm 0.03201$ & $0.128773 \pm 0.006697$ \\
        4 & Full mixture-level & $0.78961 \pm 0.03686$ & $0.115344 \pm 0.007584$ \\
        \bottomrule
    \end{tabular}
\end{table}

\begin{table}[H]
    \centering
    \small
    \setlength{\tabcolsep}{4pt}
    \caption{Individual training seeds at maximum depth 8. Each entry is the mean $\pm$ sample standard deviation across the five fixed test folds.}
    \label{tab:depth_seed_8}
    \begin{tabular}{llcc}
        \toprule
        Seed & Features & Pearson & RMSE \\
        \midrule
        0 & Absolute differences & $0.76953 \pm 0.04159$ & $0.119756 \pm 0.006640$ \\
        0 & Sums and products & $0.73946 \pm 0.02947$ & $0.128588 \pm 0.006285$ \\
        0 & Full mixture-level & $0.78371 \pm 0.03554$ & $0.116839 \pm 0.007528$ \\
        1 & Absolute differences & $0.76908 \pm 0.04183$ & $0.120004 \pm 0.007274$ \\
        1 & Sums and products & $0.73778 \pm 0.03073$ & $0.128760 \pm 0.007613$ \\
        1 & Full mixture-level & $0.78317 \pm 0.03357$ & $0.116954 \pm 0.007010$ \\
        2 & Absolute differences & $0.77096 \pm 0.04355$ & $0.119381 \pm 0.007220$ \\
        2 & Sums and products & $0.74197 \pm 0.03132$ & $0.128344 \pm 0.006442$ \\
        2 & Full mixture-level & $0.78714 \pm 0.03356$ & $0.116092 \pm 0.006964$ \\
        3 & Absolute differences & $0.76885 \pm 0.04531$ & $0.119979 \pm 0.007981$ \\
        3 & Sums and products & $0.73768 \pm 0.03114$ & $0.128808 \pm 0.007232$ \\
        3 & Full mixture-level & $0.78591 \pm 0.03445$ & $0.116232 \pm 0.006968$ \\
        4 & Absolute differences & $0.77078 \pm 0.04530$ & $0.119445 \pm 0.007540$ \\
        4 & Sums and products & $0.73997 \pm 0.02532$ & $0.128518 \pm 0.007706$ \\
        4 & Full mixture-level & $0.78506 \pm 0.03489$ & $0.116504 \pm 0.006854$ \\
        \bottomrule
    \end{tabular}
\end{table}

\begin{table}[H]
    \centering
    \small
    \setlength{\tabcolsep}{4pt}
    \caption{Individual training seeds at maximum depth 10. Each entry is the mean $\pm$ sample standard deviation across the five fixed test folds.}
    \label{tab:depth_seed_10}
    \begin{tabular}{llcc}
        \toprule
        Seed & Features & Pearson & RMSE \\
        \midrule
        0 & Absolute differences & $0.76925 \pm 0.04165$ & $0.119993 \pm 0.007053$ \\
        0 & Sums and products & $0.74445 \pm 0.02502$ & $0.127983 \pm 0.006588$ \\
        0 & Full mixture-level & $0.78204 \pm 0.03447$ & $0.117148 \pm 0.007354$ \\
        1 & Absolute differences & $0.77011 \pm 0.04153$ & $0.119685 \pm 0.006721$ \\
        1 & Sums and products & $0.73997 \pm 0.02910$ & $0.128823 \pm 0.007557$ \\
        1 & Full mixture-level & $0.78223 \pm 0.03331$ & $0.117173 \pm 0.006501$ \\
        2 & Absolute differences & $0.76737 \pm 0.04308$ & $0.120276 \pm 0.006870$ \\
        2 & Sums and products & $0.73861 \pm 0.02345$ & $0.129041 \pm 0.006923$ \\
        2 & Full mixture-level & $0.78469 \pm 0.03165$ & $0.116596 \pm 0.006588$ \\
        3 & Absolute differences & $0.76713 \pm 0.04698$ & $0.120250 \pm 0.007764$ \\
        3 & Sums and products & $0.73564 \pm 0.03048$ & $0.129493 \pm 0.008082$ \\
        3 & Full mixture-level & $0.78441 \pm 0.03242$ & $0.116662 \pm 0.006835$ \\
        4 & Absolute differences & $0.76977 \pm 0.04484$ & $0.119533 \pm 0.006809$ \\
        4 & Sums and products & $0.74025 \pm 0.02851$ & $0.128792 \pm 0.008276$ \\
        4 & Full mixture-level & $0.78437 \pm 0.03441$ & $0.116657 \pm 0.006953$ \\
        \bottomrule
    \end{tabular}
\end{table}

\begin{table}[H]
    \centering
    \small
    \setlength{\tabcolsep}{4pt}
    \caption{Individual training seeds at maximum depth 20. Each entry is the mean $\pm$ sample standard deviation across the five fixed test folds.}
    \label{tab:depth_seed_20}
    \begin{tabular}{llcc}
        \toprule
        Seed & Features & Pearson & RMSE \\
        \midrule
        0 & Absolute differences & $0.76904 \pm 0.04354$ & $0.119871 \pm 0.007267$ \\
        0 & Sums and products & $0.74388 \pm 0.02494$ & $0.128455 \pm 0.007479$ \\
        0 & Full mixture-level & $0.78315 \pm 0.03641$ & $0.116961 \pm 0.007196$ \\
        1 & Absolute differences & $0.77032 \pm 0.04001$ & $0.119687 \pm 0.006430$ \\
        1 & Sums and products & $0.73737 \pm 0.02757$ & $0.129347 \pm 0.007678$ \\
        1 & Full mixture-level & $0.78114 \pm 0.03342$ & $0.117498 \pm 0.006653$ \\
        2 & Absolute differences & $0.76884 \pm 0.04235$ & $0.119888 \pm 0.006765$ \\
        2 & Sums and products & $0.73582 \pm 0.02504$ & $0.129869 \pm 0.007754$ \\
        2 & Full mixture-level & $0.78359 \pm 0.03025$ & $0.117014 \pm 0.006102$ \\
        3 & Absolute differences & $0.76570 \pm 0.04680$ & $0.120550 \pm 0.007768$ \\
        3 & Sums and products & $0.73602 \pm 0.03207$ & $0.129661 \pm 0.007969$ \\
        3 & Full mixture-level & $0.78522 \pm 0.03291$ & $0.116545 \pm 0.006797$ \\
        4 & Absolute differences & $0.77097 \pm 0.04515$ & $0.119354 \pm 0.007002$ \\
        4 & Sums and products & $0.74004 \pm 0.02601$ & $0.128986 \pm 0.007826$ \\
        4 & Full mixture-level & $0.78424 \pm 0.03420$ & $0.116791 \pm 0.006587$ \\
        \bottomrule
    \end{tabular}
\end{table}

\begin{table}[H]
    \centering
    \small
    \setlength{\tabcolsep}{4pt}
    \caption{Individual training seeds at maximum depths 50, 100, 300, and 1000. All four depth limits give identical test predictions for each matched configuration, so their results are reported together. Each entry is the mean $\pm$ sample standard deviation across the five fixed test folds.}
    \label{tab:depth_seed_plateau}
    \begin{tabular}{llcc}
        \toprule
        Seed & Features & Pearson & RMSE \\
        \midrule
        0 & Absolute differences & $0.76902 \pm 0.04354$ & $0.119872 \pm 0.007271$ \\
        0 & Sums and products & $0.74388 \pm 0.02495$ & $0.128460 \pm 0.007546$ \\
        0 & Full mixture-level & $0.78309 \pm 0.03651$ & $0.117009 \pm 0.007220$ \\
        1 & Absolute differences & $0.77030 \pm 0.03985$ & $0.119696 \pm 0.006409$ \\
        1 & Sums and products & $0.73730 \pm 0.02722$ & $0.129318 \pm 0.007534$ \\
        1 & Full mixture-level & $0.78118 \pm 0.03341$ & $0.117488 \pm 0.006662$ \\
        2 & Absolute differences & $0.76881 \pm 0.04238$ & $0.119890 \pm 0.006764$ \\
        2 & Sums and products & $0.73703 \pm 0.02383$ & $0.129635 \pm 0.007826$ \\
        2 & Full mixture-level & $0.78367 \pm 0.03031$ & $0.116973 \pm 0.006137$ \\
        3 & Absolute differences & $0.76581 \pm 0.04678$ & $0.120514 \pm 0.007817$ \\
        3 & Sums and products & $0.73602 \pm 0.03173$ & $0.129555 \pm 0.007901$ \\
        3 & Full mixture-level & $0.78526 \pm 0.03288$ & $0.116533 \pm 0.006776$ \\
        4 & Absolute differences & $0.77097 \pm 0.04516$ & $0.119358 \pm 0.006992$ \\
        4 & Sums and products & $0.73978 \pm 0.02677$ & $0.129211 \pm 0.008076$ \\
        4 & Full mixture-level & $0.78400 \pm 0.03428$ & $0.116793 \pm 0.006616$ \\
        \bottomrule
    \end{tabular}
\end{table}

\section{Baseline Fidelity and Reproducibility}
\label{app:baseline_fidelity}

The reference implementation is the chemcognition-lab/pom-mix repository at commit
\texttt{2558557a}\allowbreak\texttt{c4793ce9}\allowbreak\texttt{82d0bd26}\allowbreak\texttt{cfa79cdf}\allowbreak\texttt{4dfbbb6f}.
The audited data files, feature archives, mixture definitions, data loader, split generator, five split files, and original baseline script match this version. The local PNA utility differs in optional imports, but its aggregation outputs match the pinned implementation exactly on the benchmark inputs. We therefore distinguish verified numerical fidelity from byte-for-byte identity of every local source file.

The frozen POM archive has shape $(760,43,196)$, and embedding rows were checked against mixture identifiers. Whole padding rows with value $-999$ are excluded before aggregation. Molecular variance and standard deviation use population definitions ($\mathrm{ddof}=0$); the reported variation across folds uses sample standard deviation ($\mathrm{ddof}=1$). Targets retain the released dissimilarity convention, with lower values indicating greater similarity; no label or prediction inversion is applied.

The five original partitions contain 605 training, 87 validation, and 173 test rows each. Seed-zero split regeneration reproduces the released partitions. The outer split is stratified by seven source labels, including study subdatasets; the inner validation split is not stratified. Rows are disjoint within each fold, but mixture identities and some repeated unordered pairs occur across partitions. These random-split results therefore do not establish generalization to entirely unseen mixtures. We retain the original partitions rather than remove duplicates or introduce new splits in this reproduction.

The original RDKit and POM XGBoost reproductions retain independent PNA aggregation, the original interleaved pair flattening, maximum depth 1000, and training-only pair-swap augmentation. Our reproductions explicitly fix training seed 0, which the released script leaves unspecified. The depth-six POM control changes only the depth limit relative to this fixed-seed reproduction. The explicit comparison models use symmetric features and no swap augmentation. All XGBoost models stop on validation MSE with patience 25; computing validation Pearson does not change this stopping criterion. Explicit-feature KRR instead selects parameters by validation Pearson, with standardization and target centering fitted using training data only.

Historical POM and RDKit baseline Pearson means of 0.4970 and 0.4847 come from older local result artifacts. The pinned-protocol reproductions give 0.5087 and 0.4908, respectively. These differences reflect different result sources, not improvements due to the new pair representation. The full training provenance of the older artifacts cannot be established from the current script alone. The POM reproduction matches a separate local output folder whose name ends in \texttt{\_k10}; the script nevertheless loads the five-fold \texttt{random\_cv} partitions.

CheMix and POMMix results are recalculated from released prediction files whose contents match the pinned repository and whose rows and labels were checked against the test partitions. We do not retrain these models or independently regenerate predictions from checkpoints. Snitz results likewise use checked existing prediction artifacts. All reported fold standard deviations use the same sample convention.

The new depth experiments retain commands, dependency versions, code snapshots and hashes, input and split hashes, training seeds, models, predictions, learning curves, and selected iterations. Saved models are reloaded to verify their predictions, and metrics are recomputed from the saved prediction files. The original manuscript and prior result artifacts are protected by file-list and SHA-256 checks.

\section{Supplementary Dense-Grid Depth Analysis}
\label{app:dense_depth}

The dense grid contains depths 6, 8, 10, 12, 16, 24, and 32. Every depth uses five training seeds, five fixed folds, and three representations. Depths 6, 8, and 10 are reused after verification; missing configurations at depths 12, 16, 24, and 32 are completed with the same protocol. The wider-grid results are retained, giving 900 unique configurations over 12 depth limits. The supplementary depths were requested after partial wider-grid results were inspected. No evaluated depth is omitted from the combined summary, and test results are not used to replace the primary depth-six model.

Overall tables average metrics over seeds within each fold before reporting five-fold means and sample standard deviations ($\mathrm{ddof}=1$). Individual-seed tables instead summarize that seed's five folds. The 25 seed--fold evaluations are not treated as independent datasets. No seed-ensemble predictions are used. Models, predictions, learning curves, individual-seed summaries, paired differences, and actual-tree-depth diagnostics are retained with the experiment artifacts.

\begin{table}[H]
    \centering
    \small
    \setlength{\tabcolsep}{4pt}
    \caption{Test metrics on the dense grid. Entries are mean $\pm$ sample standard deviation across folds after within-fold averaging over seeds.}
    \label{tab:supplement_dense}
    \begin{tabular}{llcc}
        \toprule
        Depth & Features & Pearson & RMSE \\
        \midrule
        6 & Absolute differences & $0.77163 \pm 0.04323$ & $0.119305 \pm 0.007107$ \\
        6 & Sums and products & $0.73769 \pm 0.03205$ & $0.128635 \pm 0.007150$ \\
        6 & Full mixture-level & $0.78769 \pm 0.03610$ & $0.115795 \pm 0.007301$ \\
        8 & Absolute differences & $0.76984 \pm 0.04337$ & $0.119713 \pm 0.007295$ \\
        8 & Sums and products & $0.73937 \pm 0.02927$ & $0.128603 \pm 0.007030$ \\
        8 & Full mixture-level & $0.78500 \pm 0.03431$ & $0.116524 \pm 0.007044$ \\
        10 & Absolute differences & $0.76873 \pm 0.04354$ & $0.119947 \pm 0.007019$ \\
        10 & Sums and products & $0.73978 \pm 0.02685$ & $0.128826 \pm 0.007447$ \\
        10 & Full mixture-level & $0.78355 \pm 0.03319$ & $0.116847 \pm 0.006831$ \\
        12 & Absolute differences & $0.76823 \pm 0.04360$ & $0.120035 \pm 0.007178$ \\
        12 & Sums and products & $0.73932 \pm 0.02593$ & $0.129036 \pm 0.007695$ \\
        12 & Full mixture-level & $0.78379 \pm 0.03361$ & $0.116812 \pm 0.006796$ \\
        16 & Absolute differences & $0.76833 \pm 0.04331$ & $0.119970 \pm 0.006947$ \\
        16 & Sums and products & $0.73935 \pm 0.02571$ & $0.129137 \pm 0.007768$ \\
        16 & Full mixture-level & $0.78391 \pm 0.03374$ & $0.116840 \pm 0.006771$ \\
        24 & Absolute differences & $0.76898 \pm 0.04344$ & $0.119866 \pm 0.007020$ \\
        24 & Sums and products & $0.73889 \pm 0.02677$ & $0.129248 \pm 0.007732$ \\
        24 & Full mixture-level & $0.78345 \pm 0.03343$ & $0.116954 \pm 0.006656$ \\
        32 & Absolute differences & $0.76898 \pm 0.04344$ & $0.119866 \pm 0.007020$ \\
        32 & Sums and products & $0.73879 \pm 0.02664$ & $0.129237 \pm 0.007735$ \\
        32 & Full mixture-level & $0.78344 \pm 0.03341$ & $0.116959 \pm 0.006660$ \\
        \bottomrule
    \end{tabular}
\end{table}

\begin{table}[H]
    \centering
    \small
    \setlength{\tabcolsep}{4pt}
    \caption{Test metrics at the remaining evaluated depths. Entries are mean $\pm$ sample standard deviation across folds after within-fold averaging over seeds.}
    \label{tab:supplement_wider}
    \begin{tabular}{llcc}
        \toprule
        Depth & Features & Pearson & RMSE \\
        \midrule
        20 & Absolute differences & $0.76897 \pm 0.04347$ & $0.119870 \pm 0.007017$ \\
        20 & Sums and products & $0.73863 \pm 0.02696$ & $0.129263 \pm 0.007712$ \\
        20 & Full mixture-level & $0.78347 \pm 0.03337$ & $0.116962 \pm 0.006648$ \\
        50 & Absolute differences & $0.76898 \pm 0.04344$ & $0.119866 \pm 0.007020$ \\
        50 & Sums and products & $0.73880 \pm 0.02665$ & $0.129236 \pm 0.007736$ \\
        50 & Full mixture-level & $0.78344 \pm 0.03341$ & $0.116959 \pm 0.006660$ \\
        100 & Absolute differences & $0.76898 \pm 0.04344$ & $0.119866 \pm 0.007020$ \\
        100 & Sums and products & $0.73880 \pm 0.02665$ & $0.129236 \pm 0.007736$ \\
        100 & Full mixture-level & $0.78344 \pm 0.03341$ & $0.116959 \pm 0.006660$ \\
        300 & Absolute differences & $0.76898 \pm 0.04344$ & $0.119866 \pm 0.007020$ \\
        300 & Sums and products & $0.73880 \pm 0.02665$ & $0.129236 \pm 0.007736$ \\
        300 & Full mixture-level & $0.78344 \pm 0.03341$ & $0.116959 \pm 0.006660$ \\
        1000 & Absolute differences & $0.76898 \pm 0.04344$ & $0.119866 \pm 0.007020$ \\
        1000 & Sums and products & $0.73880 \pm 0.02665$ & $0.129236 \pm 0.007736$ \\
        1000 & Full mixture-level & $0.78344 \pm 0.03341$ & $0.116959 \pm 0.006660$ \\
        \bottomrule
    \end{tabular}
\end{table}



\begin{table}[H]
    \centering
    \small
    \setlength{\tabcolsep}{4pt}
    \caption{Paired gaps on the dense grid. Pearson is absdiff minus sum+product; RMSE is sum+product minus absdiff. Positive gaps favor absdiff. The last column subtracts the depth-six mean Pearson gap.}
    \label{tab:dense_paired_gaps}
    \begin{tabular}{lccc}
        \toprule
        Depth & Pearson gap & RMSE gap & Gap change \\
        \midrule
        6 & $0.03394 \pm 0.01628$ & $0.009329 \pm 0.005569$ & $+0.00000$ \\
        8 & $0.03047 \pm 0.01714$ & $0.008890 \pm 0.005861$ & $-0.00347$ \\
        10 & $0.02894 \pm 0.02159$ & $0.008879 \pm 0.006609$ & $-0.00500$ \\
        12 & $0.02892 \pm 0.02271$ & $0.009001 \pm 0.006890$ & $-0.00502$ \\
        16 & $0.02898 \pm 0.02325$ & $0.009167 \pm 0.007035$ & $-0.00496$ \\
        24 & $0.03010 \pm 0.02204$ & $0.009382 \pm 0.006849$ & $-0.00384$ \\
        32 & $0.03019 \pm 0.02218$ & $0.009371 \pm 0.006857$ & $-0.00375$ \\
        \bottomrule
    \end{tabular}
\end{table}

\begin{table}[H]
    \centering
    \small
    \setlength{\tabcolsep}{4pt}
    \caption{Pearson gap by fixed fold, after averaging matched-seed gaps within each fold. Positive values favor absdiff.}
    \label{tab:dense_folds_pearson_gap}
    \begin{tabular}{lccccc}
        \toprule
        Depth & cv0 & cv1 & cv2 & cv3 & cv4 \\
        \midrule
        6 & $+0.045112$ & $+0.010859$ & $+0.034788$ & $+0.052528$ & $+0.026406$ \\
        8 & $+0.040733$ & $+0.012702$ & $+0.030148$ & $+0.053408$ & $+0.015365$ \\
        10 & $+0.041745$ & $+0.003703$ & $+0.026984$ & $+0.057997$ & $+0.014277$ \\
        12 & $+0.043292$ & $+0.002722$ & $+0.023190$ & $+0.059861$ & $+0.015522$ \\
        16 & $+0.044247$ & $+0.000868$ & $+0.024156$ & $+0.059724$ & $+0.015922$ \\
        24 & $+0.045689$ & $+0.002887$ & $+0.024778$ & $+0.058419$ & $+0.018703$ \\
        32 & $+0.046062$ & $+0.002877$ & $+0.024656$ & $+0.058613$ & $+0.018733$ \\
        \bottomrule
    \end{tabular}
\end{table}

\begin{table}[H]
    \centering
    \small
    \setlength{\tabcolsep}{4pt}
    \caption{RMSE gap by fixed fold, after averaging matched-seed gaps within each fold. Positive values favor absdiff.}
    \label{tab:dense_folds_rmse_gap}
    \begin{tabular}{lccccc}
        \toprule
        Depth & cv0 & cv1 & cv2 & cv3 & cv4 \\
        \midrule
        6 & $+0.013110$ & $+0.003359$ & $+0.007501$ & $+0.016937$ & $+0.005740$ \\
        8 & $+0.012397$ & $+0.003873$ & $+0.006955$ & $+0.017331$ & $+0.003894$ \\
        10 & $+0.013074$ & $+0.002292$ & $+0.006845$ & $+0.018156$ & $+0.004028$ \\
        12 & $+0.013446$ & $+0.002084$ & $+0.006414$ & $+0.018701$ & $+0.004361$ \\
        16 & $+0.014043$ & $+0.001803$ & $+0.006765$ & $+0.018762$ & $+0.004462$ \\
        24 & $+0.014276$ & $+0.002214$ & $+0.006968$ & $+0.018636$ & $+0.004817$ \\
        32 & $+0.014242$ & $+0.002181$ & $+0.006956$ & $+0.018652$ & $+0.004823$ \\
        \bottomrule
    \end{tabular}
\end{table}

\begin{table}[H]
    \centering
    \small
    \setlength{\tabcolsep}{4pt}
    \caption{Dense-grid training diagnostics. Training RMSE and tree counts use within-fold seed averaging, followed by fold mean and sample standard deviation. Actual depth is the range of per-model maximum selected-tree depths across 25 configurations.}
    \label{tab:dense_training}
    \begin{tabular}{llccc}
        \toprule
        Limit & Features & Train RMSE & Actual depth & Selected trees \\
        \midrule
        6 & Absolute differences & $0.03086 \pm 0.00697$ & 6--6 & $552.6 \pm 216.0$ \\
        6 & Sums and products & $0.01533 \pm 0.00359$ & 6--6 & $756.8 \pm 152.1$ \\
        6 & Full mixture-level & $0.01716 \pm 0.00483$ & 6--6 & $578.6 \pm 155.9$ \\
        8 & Absolute differences & $0.02737 \pm 0.00500$ & 8--8 & $578.6 \pm 250.4$ \\
        8 & Sums and products & $0.01030 \pm 0.00267$ & 8--8 & $767.1 \pm 140.1$ \\
        8 & Full mixture-level & $0.01220 \pm 0.00424$ & 8--8 & $613.8 \pm 190.7$ \\
        10 & Absolute differences & $0.02698 \pm 0.00546$ & 10--10 & $567.2 \pm 239.3$ \\
        10 & Sums and products & $0.00959 \pm 0.00285$ & 10--10 & $728.3 \pm 154.5$ \\
        10 & Full mixture-level & $0.01068 \pm 0.00339$ & 10--10 & $603.4 \pm 128.8$ \\
        12 & Absolute differences & $0.02652 \pm 0.00500$ & 12--12 & $568.1 \pm 221.3$ \\
        12 & Sums and products & $0.00910 \pm 0.00334$ & 12--12 & $715.3 \pm 137.5$ \\
        12 & Full mixture-level & $0.01078 \pm 0.00368$ & 12--12 & $577.7 \pm 131.3$ \\
        16 & Absolute differences & $0.02612 \pm 0.00494$ & 16--16 & $576.8 \pm 198.2$ \\
        16 & Sums and products & $0.00913 \pm 0.00286$ & 16--16 & $696.9 \pm 106.0$ \\
        16 & Full mixture-level & $0.01111 \pm 0.00429$ & 16--16 & $571.8 \pm 144.8$ \\
        24 & Absolute differences & $0.02665 \pm 0.00545$ & 18--24 & $549.9 \pm 213.7$ \\
        24 & Sums and products & $0.00932 \pm 0.00294$ & 24--24 & $664.2 \pm 115.0$ \\
        24 & Full mixture-level & $0.01100 \pm 0.00396$ & 18--24 & $555.3 \pm 131.9$ \\
        32 & Absolute differences & $0.02665 \pm 0.00545$ & 18--28 & $552.8 \pm 216.3$ \\
        32 & Sums and products & $0.00911 \pm 0.00306$ & 24--32 & $688.2 \pm 120.7$ \\
        32 & Full mixture-level & $0.01101 \pm 0.00390$ & 18--28 & $554.6 \pm 128.5$ \\
        \bottomrule
    \end{tabular}
\end{table}

\begin{table}[H]
    \centering
    \small
    \setlength{\tabcolsep}{4pt}
    \caption{Individual training seeds at maximum depth 12. Each entry is the mean $\pm$ sample standard deviation across five folds.}
    \label{tab:dense_seed_12}
    \begin{tabular}{llcc}
        \toprule
        Seed & Features & Pearson & RMSE \\
        \midrule
        0 & Absolute differences & $0.76863 \pm 0.04343$ & $0.119907 \pm 0.007271$ \\
        0 & Sums and products & $0.74243 \pm 0.02210$ & $0.128568 \pm 0.007478$ \\
        0 & Full mixture-level & $0.78377 \pm 0.03523$ & $0.116802 \pm 0.007353$ \\
        1 & Absolute differences & $0.76975 \pm 0.04167$ & $0.119768 \pm 0.006895$ \\
        1 & Sums and products & $0.73912 \pm 0.02593$ & $0.129020 \pm 0.007304$ \\
        1 & Full mixture-level & $0.78180 \pm 0.03363$ & $0.117238 \pm 0.006525$ \\
        2 & Absolute differences & $0.76846 \pm 0.04267$ & $0.119976 \pm 0.006839$ \\
        2 & Sums and products & $0.73680 \pm 0.02371$ & $0.129503 \pm 0.007464$ \\
        2 & Full mixture-level & $0.78419 \pm 0.03095$ & $0.116789 \pm 0.006437$ \\
        3 & Absolute differences & $0.76531 \pm 0.04678$ & $0.120697 \pm 0.007950$ \\
        3 & Sums and products & $0.73627 \pm 0.03210$ & $0.129567 \pm 0.007991$ \\
        3 & Full mixture-level & $0.78636 \pm 0.03270$ & $0.116315 \pm 0.006913$ \\
        4 & Absolute differences & $0.76903 \pm 0.04389$ & $0.119828 \pm 0.007030$ \\
        4 & Sums and products & $0.74196 \pm 0.02689$ & $0.128523 \pm 0.008411$ \\
        4 & Full mixture-level & $0.78283 \pm 0.03576$ & $0.116918 \pm 0.006795$ \\
        \bottomrule
    \end{tabular}
\end{table}

\begin{table}[H]
    \centering
    \small
    \setlength{\tabcolsep}{4pt}
    \caption{Individual training seeds at maximum depth 16. Each entry is the mean $\pm$ sample standard deviation across five folds.}
    \label{tab:dense_seed_16}
    \begin{tabular}{llcc}
        \toprule
        Seed & Features & Pearson & RMSE \\
        \midrule
        0 & Absolute differences & $0.76888 \pm 0.04332$ & $0.119904 \pm 0.007256$ \\
        0 & Sums and products & $0.74304 \pm 0.02346$ & $0.128681 \pm 0.007564$ \\
        0 & Full mixture-level & $0.78423 \pm 0.03660$ & $0.116708 \pm 0.007404$ \\
        1 & Absolute differences & $0.77029 \pm 0.04021$ & $0.119567 \pm 0.006354$ \\
        1 & Sums and products & $0.73743 \pm 0.02734$ & $0.129399 \pm 0.007691$ \\
        1 & Full mixture-level & $0.78274 \pm 0.03299$ & $0.117059 \pm 0.006445$ \\
        2 & Absolute differences & $0.76789 \pm 0.04224$ & $0.120138 \pm 0.006611$ \\
        2 & Sums and products & $0.73901 \pm 0.02267$ & $0.129334 \pm 0.007877$ \\
        2 & Full mixture-level & $0.78380 \pm 0.03077$ & $0.117001 \pm 0.006265$ \\
        3 & Absolute differences & $0.76541 \pm 0.04699$ & $0.120546 \pm 0.007743$ \\
        3 & Sums and products & $0.73734 \pm 0.03132$ & $0.129304 \pm 0.008058$ \\
        3 & Full mixture-level & $0.78489 \pm 0.03354$ & $0.116606 \pm 0.006937$ \\
        4 & Absolute differences & $0.76919 \pm 0.04437$ & $0.119694 \pm 0.006894$ \\
        4 & Sums and products & $0.73991 \pm 0.02505$ & $0.128966 \pm 0.007874$ \\
        4 & Full mixture-level & $0.78392 \pm 0.03521$ & $0.116825 \pm 0.006913$ \\
        \bottomrule
    \end{tabular}
\end{table}

\begin{table}[H]
    \centering
    \small
    \setlength{\tabcolsep}{4pt}
    \caption{Individual training seeds at maximum depth 24. Each entry is the mean $\pm$ sample standard deviation across five folds.}
    \label{tab:dense_seed_24}
    \begin{tabular}{llcc}
        \toprule
        Seed & Features & Pearson & RMSE \\
        \midrule
        0 & Absolute differences & $0.76903 \pm 0.04354$ & $0.119871 \pm 0.007274$ \\
        0 & Sums and products & $0.74409 \pm 0.02505$ & $0.128412 \pm 0.007506$ \\
        0 & Full mixture-level & $0.78309 \pm 0.03652$ & $0.117011 \pm 0.007221$ \\
        1 & Absolute differences & $0.77030 \pm 0.03987$ & $0.119695 \pm 0.006410$ \\
        1 & Sums and products & $0.73750 \pm 0.02764$ & $0.129355 \pm 0.007541$ \\
        1 & Full mixture-level & $0.78118 \pm 0.03341$ & $0.117492 \pm 0.006658$ \\
        2 & Absolute differences & $0.76882 \pm 0.04237$ & $0.119888 \pm 0.006765$ \\
        2 & Sums and products & $0.73706 \pm 0.02383$ & $0.129666 \pm 0.007804$ \\
        2 & Full mixture-level & $0.78367 \pm 0.03031$ & $0.116973 \pm 0.006137$ \\
        3 & Absolute differences & $0.76582 \pm 0.04680$ & $0.120519 \pm 0.007808$ \\
        3 & Sums and products & $0.73603 \pm 0.03198$ & $0.129608 \pm 0.007943$ \\
        3 & Full mixture-level & $0.78525 \pm 0.03289$ & $0.116535 \pm 0.006783$ \\
        4 & Absolute differences & $0.77095 \pm 0.04514$ & $0.119357 \pm 0.006994$ \\
        4 & Sums and products & $0.73977 \pm 0.02666$ & $0.129200 \pm 0.008087$ \\
        4 & Full mixture-level & $0.78407 \pm 0.03432$ & $0.116762 \pm 0.006590$ \\
        \bottomrule
    \end{tabular}
\end{table}

\begin{table}[H]
    \centering
    \small
    \setlength{\tabcolsep}{4pt}
    \caption{Individual training seeds at maximum depth 32. Each entry is the mean $\pm$ sample standard deviation across five folds.}
    \label{tab:dense_seed_32}
    \begin{tabular}{llcc}
        \toprule
        Seed & Features & Pearson & RMSE \\
        \midrule
        0 & Absolute differences & $0.76902 \pm 0.04354$ & $0.119872 \pm 0.007271$ \\
        0 & Sums and products & $0.74388 \pm 0.02495$ & $0.128460 \pm 0.007546$ \\
        0 & Full mixture-level & $0.78309 \pm 0.03651$ & $0.117009 \pm 0.007220$ \\
        1 & Absolute differences & $0.77030 \pm 0.03985$ & $0.119696 \pm 0.006409$ \\
        1 & Sums and products & $0.73726 \pm 0.02720$ & $0.129322 \pm 0.007526$ \\
        1 & Full mixture-level & $0.78118 \pm 0.03341$ & $0.117488 \pm 0.006662$ \\
        2 & Absolute differences & $0.76881 \pm 0.04238$ & $0.119890 \pm 0.006764$ \\
        2 & Sums and products & $0.73702 \pm 0.02382$ & $0.129636 \pm 0.007826$ \\
        2 & Full mixture-level & $0.78367 \pm 0.03031$ & $0.116973 \pm 0.006137$ \\
        3 & Absolute differences & $0.76581 \pm 0.04678$ & $0.120514 \pm 0.007817$ \\
        3 & Sums and products & $0.73603 \pm 0.03174$ & $0.129554 \pm 0.007905$ \\
        3 & Full mixture-level & $0.78526 \pm 0.03288$ & $0.116533 \pm 0.006776$ \\
        4 & Absolute differences & $0.77097 \pm 0.04516$ & $0.119358 \pm 0.006992$ \\
        4 & Sums and products & $0.73978 \pm 0.02677$ & $0.129211 \pm 0.008076$ \\
        4 & Full mixture-level & $0.78400 \pm 0.03428$ & $0.116793 \pm 0.006616$ \\
        \bottomrule
    \end{tabular}
\end{table}

\section{Diagnostic Comparisons of KRR and XGBoost}
\label{app:krr_diagnostics}

We conducted follow-up diagnostics to examine the gap between RBF KRR and XGBoost. All experiments used the full 2352-dimensional mixture-level comparison representation and the same five random splits, with 605 training, 87 validation, and 173 test samples per fold. No cross-mixture interaction features or pair-swap augmentation were used. These variants were investigated after inspecting the primary results and are reported as exploratory diagnostics rather than replacements for the primary models.

\subsection{Preprocessing and Model Selection}

For the standard KRR variants, each feature was standardized using the training-fold mean and standard deviation. For the quantile variants, this preprocessing was replaced by a coordinate-wise quantile transformation fitted to the raw training features, using 200 quantiles, standard normal output marginals, and random seed 0. The fitted transformation was then applied unchanged to validation and test features. KRR labels were centered using the training-fold mean, which was added back to predictions.

We compared linear, RBF, and Laplacian kernels. The nonlinear kernels were
\[
k_{\mathrm{RBF}}(u,v)
=
\exp\left(-\frac{\|u-v\|_2^2}{\sigma^2}\right),
\qquad
k_{\mathrm{Lap}}(u,v)
=
\exp\left(-\frac{\|u-v\|_1}{\sigma}\right).
\]
The regularization grid for the nonlinear kernels was
\[
\alpha \in
\{10^{-4},10^{-3},10^{-2},10^{-1},1,10,100,1000\}.
\]
Parameters were selected by validation Pearson correlation within each fold.

The original RBF comparison used bandwidths
\[
\sigma \in \{0.25,0.5,1,2,4\}\sqrt{d},
\qquad d=2352.
\]
An additional RBF check instead used
\[
\sigma \in \{0.25,0.5,1,2,4,8,16\}m_2,
\]
where $m_2$ is the median Euclidean distance between distinct training samples. The quantile-transformed and rotated RBF variants used the factors $\{0.25,0.5,1,2,4,8\}$ with the median distance computed after preprocessing. Laplacian variants used the same six factors with the median training Manhattan distance. These grids use different bandwidth scales and should not be interpreted as a strictly nested expansion of the original grid.

For the rotation diagnostic, we generated one orthogonal matrix by QR decomposition of a Gaussian random matrix with random seed 0. The same matrix was applied to training, validation, and test features after training-fold standardization, and was reused across folds. No further standardization was performed after rotation.

The XGBoost diagnostic controls used training seed 0, maximum depth 6, at most 1000 trees, learning rate 0.01, row and column subsampling rates of 0.8, and early stopping after 25 rounds without improvement in validation MSE. The unrotated diagnostic control used standardized inputs, whereas the primary XGBoost model used the original comparison features.

\subsection{Results and Interpretation}

Table~\ref{tab:krr_diagnostics} reports the main diagnostic comparisons. Laplacian KRR with quantile transformation improves mean Pearson correlation over the original RBF KRR by $0.0285$ and reduces mean RMSE by $0.0066$. Pearson correlation improves in all five folds. Compared with the standardized-input XGBoost control, its mean Pearson correlation is lower by $0.0077$ and its mean RMSE is higher by $0.0020$. It has higher Pearson correlation in two folds and lower correlation in three folds. These results indicate a smaller observed gap, but do not establish statistical equivalence.

\begin{table}[H]
    \centering
    \small
    \caption{Follow-up diagnostics using the full mixture-level comparison representation. Values are mean $\pm$ sample standard deviation across the same five folds ($\mathrm{ddof}=1$). Standard KRR inputs are standardized within the training fold; quantile variants replace standardization with a training-fitted quantile transformation. The primary XGBoost result is included for reference.}
    \label{tab:krr_diagnostics}
    \begin{tabular}{@{}p{0.46\linewidth}cc@{}}
        \toprule
        Model / preprocessing & Pearson & RMSE \\
        \midrule
        XGBoost, primary inputs
        & $0.7861 \pm 0.0359$ & $0.1161 \pm 0.0070$ \\
        XGBoost, standardized
        & $0.7866 \pm 0.0370$ & $0.1159 \pm 0.0071$ \\
        XGBoost, standardized and rotated
        & $0.7561 \pm 0.0316$ & $0.1243 \pm 0.0060$ \\
        \midrule
        KRR, linear
        & $0.7364 \pm 0.0386$ & $0.1267 \pm 0.0047$ \\
        KRR, original RBF grid
        & $0.7504 \pm 0.0343$ & $0.1245 \pm 0.0059$ \\
        KRR, median-distance RBF grid
        & $0.7476 \pm 0.0386$ & $0.1250 \pm 0.0059$ \\
        KRR, RBF after rotation
        & $0.7476 \pm 0.0386$ & $0.1250 \pm 0.0059$ \\
        KRR, RBF with quantile transformation
        & $0.7623 \pm 0.0379$ & $0.1215 \pm 0.0076$ \\
        KRR, Laplacian
        & $0.7673 \pm 0.0423$ & $0.1206 \pm 0.0068$ \\
        KRR, Laplacian with quantile transformation
        & $0.7789 \pm 0.0370$ & $0.1179 \pm 0.0079$ \\
        \bottomrule
    \end{tabular}
\end{table}

\begin{figure}[H]
    \centering
    \includegraphics[width=\linewidth]{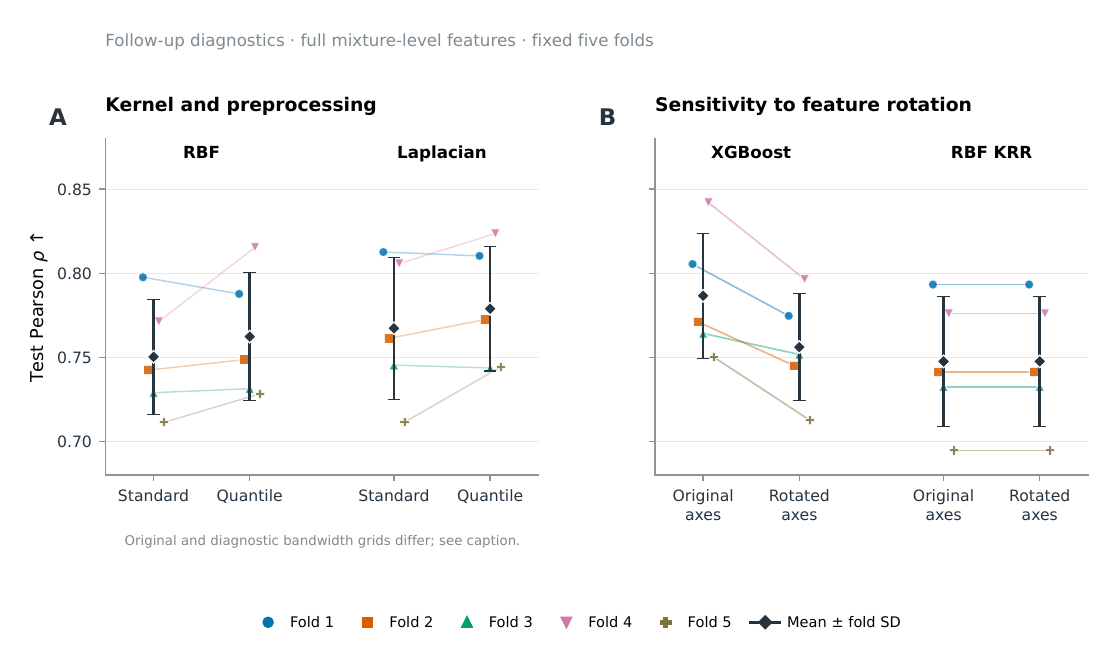}
    \caption{Follow-up diagnostics on the 2352-dimensional mixture-level representation. (A) RBF and Laplacian KRR with training-fold standardization (Standard) or a training-fitted quantile transformation to approximately normal marginals (Quantile). These variants use the bandwidth grids specified in Appendix~\ref{app:krr_diagnostics}; the original RBF grid and the quantile-transformed RBF grid differ. (B) XGBoost and RBF KRR before and after one fixed orthogonal rotation of standardized features. The XGBoost control here uses standardized inputs and is distinct from the primary unstandardized-input model. The RBF rotation comparison uses median-distance bandwidth grids; the unrotated grid additionally includes factor 16, which is not selected in any fold. Colors and marker shapes identify folds, and connecting lines pair evaluations within each fold. Black diamonds and error bars indicate means and sample standard deviations across five folds. Kernel and preprocessing changes do not isolate a unique causal mechanism, and rotation sensitivity is assessed for only one rotation.}
    \label{fig:krr_diagnostics}
\end{figure}

Figure~\ref{fig:krr_diagnostics} shows the fold-paired changes under kernel, preprocessing, and rotation diagnostics.

The median-distance RBF bandwidth check did not improve on the original RBF result. Selecting the top 50, 100, 250, 500, or 1000 features by XGBoost gain also did not improve mean test Pearson correlation over the original full-feature RBF model; the top-500 variant achieved $0.7419 \pm 0.0365$. Gain-based feature weighting produced only a small improvement, reaching a mean Pearson correlation of $0.7527$. These checks do not support a simple explanation based on the particular bandwidth grid or feature-selection procedure tested, but they do not rule out other tuning strategies or effects of irrelevant features.

Orthogonal rotation reduced XGBoost's mean Pearson correlation by $0.0305$. The rotated RBF result matched the unrotated median-distance result to numerical precision in the recorded fold metrics. This is expected because orthogonal rotation preserves Euclidean distances. The unrotated check additionally included bandwidth factor 16, which was not selected in any fold. This diagnostic supports sensitivity of XGBoost to the supplied coordinate system, but uses only one rotation.

The kernel and preprocessing changes do not isolate a unique cause of the original performance gap. Replacing the RBF kernel with the Laplacian kernel changes both the distance metric and kernel smoothness. Quantile transformation changes the marginal distributions and therefore the distances between samples. Its benefit does not by itself establish that heavy tails caused the original shortfall. Likewise, the rotation result does not establish that the target function is intrinsically nonsmooth. The supported conclusion is that KRR performance in this benchmark depends substantially on the kernel and preprocessing applied to the explicit comparison features.

All comparisons reuse the original five splits. Fold standard deviations describe variation across those splits and are not tests of equivalence between models. Because the diagnostic variants were motivated by the primary results, their performance should be confirmed under additional evaluation settings before making broader claims.